\documentclass[preprint,12pt,authoryear]{elsarticle}

\usepackage{apalike}
\usepackage{threeparttable}
\usepackage{graphicx}
\usepackage{listings}
\usepackage{float}
\usepackage{booktabs}
\usepackage{multirow}
\usepackage{algorithm} 
\usepackage{algpseudocode}  
\usepackage{amsmath} 
\usepackage{makecell}
\usepackage{subcaption}
\usepackage{caption}
\usepackage[colorlinks,
linkcolor=blue,
anchorcolor=blue,
citecolor=blue]{hyperref}

\usepackage{algpseudocode}
\usepackage{hyperref}
\usepackage{booktabs}
\newcommand{\botrule}{\bottomrule}
\usepackage{mathrsfs}
\usepackage{graphicx}
\usepackage{amsmath}
\usepackage{bm}
\usepackage{amsthm}
\usepackage{amsmath} %
\usepackage{amssymb} %
\usepackage{tabularx}
\newtheorem{definition}{Definition}
\begin{document}

\begin{frontmatter}

\title{TANGNN: a Concise, Scalable and Effective Graph Neural Networks with Top-m Attention Mechanism for  Graph Representation Learning}

\author{Jiawei E}
\ead{1092349066@qq.com}		
\author{Yinglong Zhang\corref{cor1}}
\ead{zhang\_yinglong@126.com}


\author{Xuewen Xia}
\ead{xwxia@whu.edu.cn}
\author{Xing Xu}
\ead{xxustar@qq.com}

\address{School of Physics and Information Engineering, Minnan Normal University, Zhangzhou, 363000, China}
\cortext[cor1]{Corresponding author at: School of Physics and Information Engineering, Minnan Normal University, Zhangzhou, China}
		
\begin{abstract}
In the field of deep learning, Graph Neural Networks (GNNs) and Graph Transformer models, with their outstanding performance and flexible architectural designs, have become leading technologies for processing structured data, especially graph data. Traditional GNNs often face challenges in capturing information from distant vertices effectively. In contrast, Graph Transformer models are particularly adept at managing long-distance node relationships. Despite these advantages, Graph Transformer models still encounter issues with computational and storage efficiency when scaled to large graph datasets. To address these challenges, we propose an innovative Graph Neural Network (GNN) architecture that integrates a Top-$m$ attention mechanism aggregation component and a neighborhood aggregation component, effectively enhancing the model's ability to aggregate relevant information from both local and extended neighborhoods at each layer. This method not only improves computational efficiency but also enriches the node features, facilitating a deeper analysis of complex graph structures. Additionally, to assess the effectiveness of our proposed model, we have applied it to citation sentiment prediction---a novel task previously unexplored in the GNN field. Accordingly, we constructed a dedicated citation network, ArXivNet. In this dataset, we specifically annotated the sentiment polarity of the citations (positive, neutral, negative) to enable in-depth sentiment analysis. Our approach has shown superior performance across a variety of tasks including vertex classification, link prediction, sentiment prediction, graph regression, and visualization. It outperforms existing methods in terms of effectiveness, as demonstrated by experimental results on multiple datasets. The code and ArXivNet dataset are available at \href{https://github.com/ejwww/TANGNN}{https://github.com/ejwww/TANGNN}.
\end{abstract}

		
\begin{keyword}
Graph neural network \sep 
Top-$m$ attention mechanism aggregation \sep 
Neighborhood aggregation \sep
Link prediction \sep 
Sentiment prediction
\end{keyword}
		
\end{frontmatter}
	
	
\section{Introduction}
In the field of deep learning, dealing with structured data has always been a hot topic of research. Structured data, especially graph data, plays a critical role in a variety of application scenarios, such as social networks~\citep{bib27}, citation network analysis~\citep{bib40,bib41}, recommendation systems~\citep{bib37}, knowledge graph representation~\citep{bib38}, text embedding~\cite{bib39} and bioinformatics~\citep{bib48}. However, traditional neural network architectures face numerous challenges when handling graph datasets. Against this backdrop, Graph Neural Networks (GNNs) models have become frontier technologies due to their exceptional performance and flexible architectural design.

Graph neural networks(GNN) directly utilize the topological structure of graphs by  adjacency matrices or aggregating node features, they perform well in many graph data processing tasks, such as vertex classification~\citep{bib36} and link prediction~\citep{bib35}. However, they have certain limitations when dealing with node information with farther hops. In traditional graph neural networks, the receptive field of each node is limited to its direct neighbors. In order to expand the receptive field of the model, it is necessary to increase the number of layers in the network, so that nodes can indirectly contact further neighbors. However, as network layers increase, node representations become less distinct, complicating the task of accurately labeling nodes by classifiers~\citep{bib20}. Moreover, the addition of layers leads to a decrease in training accuracy, and the emergence of oversmoothing issues~\citep{bib18,bib19,bib21}, which collectively degrade the performance of graph neural networks.

Conversely, Transformers~\cite{bib2} introduce a self-attention mechanism that allows the model to directly compute dependencies between all elements on a global scale. Although Transformers have gained significant recognition and success across a variety of domains such as machine translation~\citep{bib45}, image segmentation~\citep{bib46,bib47}, applying standard Transformers directly to graph data processing still poses challenges. Transformers are primarily designed for processing sequential data. As the input sequence lengthens, the demands for computation and storage increase quadratically, leading to a dramatic increase in resource consumption, which may reduce the efficiency of the self-attention mechanism of Transformers in processing long sequences~\citep{bib22,bib23}. Existing graph Transformers~\citep{bib42} have introduced positional and structural encodings to enhance the model's perception of node positions and structures within a graph, enabling the model to understand not only the features of nodes and edges but also their relative importance and relationships within the entire graph structure. For example, Graphormer~\citep{bib11} introduced centrality encoding, spatial encoding, and edge encoding, which help it better handle the graph data structure. TransGNN~\citep{bib8} introduced multiple different positional encodings to help the model capture the structural information of the graph. However, the introduction of these encodings also brings significant computational and storage overhead. This is primarily reflected in additional preprocessing steps, increased model complexity and memory requirements, and potentially extended training times. Especially when dealing with large-scale graph data, these overheads can significantly impact the scalability and practicality of the model.

In this paper, we propose a concise, scalable and effective GNN, TANGNN (\textbf{T}op-$m$ \textbf{A}ttention and \textbf{N}eighborhood Aggregation \textbf{GNN}). It not only effectively integrates local and global information, maintains efficient computation but also significantly enhances the model's expressive power and performance, making it well-suited for large-scale graph data. The main contributions of this paper are as follows:

\begin{itemize}
	\item We propose a new GNN framework named TANGNN. It introduces two distinct aggregation components: a Top-$m$ attention mechanism aggregation component and a neighborhood aggregation component. which effectively enhancing the model's ability to aggregate relevant information from both local and extended neighborhoods at each layer. It solves the problem of the narrow receptive field of traditional GNNs, while not bringing high computational overheads like transformers.
	\item In the Top-$m$ attention mechanism aggregation component, an auxiliary vector $a$ is introduced, and the Top-$m$ most relevant node features are selected based on the cosine similarity between the nodes and the auxiliary vector $a$. This ensures that the model focuses on the most informative features, thereby enhancing processing efficiency and the model's performance.
	\item Both the Top-$m$ attention mechanism aggregation component and the neighborhood aggregation component utilize sampling strategies, significantly reducing the memory and computational demands during the model's training and inference processes. This makes the model easily applicable to medium and large-scale graphs.
	\item To support our innovative application in the field of Graph Neural Networks(GNNs)---citation sentiment prediction, we have specifically constructed the ArXivNet dataset. To our knowledge, this is the first time citation sentiment prediction has been introduced into GNN research.
	\item Extensive experiments have been conducted on multiple real datasets, comparing our approach with several representative algorithms. The experimental results demonstrate that our method achieves superior performance on graph datasets and shows distinct advantages in tasks like node classification, link prediction, sentiment prediction, graph regression and visualization.
\end{itemize}

	\section{Related work}

Recently, a variety of new approaches have been developed in the field of graph representation learning, which are designed to increase the expressive capabilities of graph structures.

Methods based on graph neural networks: Graph Neural Networks (GNNs) are a popular research direction in the current deep learning field, mainly used for processing graph-structured data. Globally, research in this field is rapidly developing and showing a rich array of application prospects. In recent years, significant progress has been made in the study of GNNs and their variants. Methods like GCN~\citep{bib1} and SGCN~\citep{bib5} aggregate neighbor node information through multiple convolution layers. While these methods are effective at capturing the local structure around nodes, they struggle to adequately handle global structural aspects. Some approaches concentrate on enhancing techniques for neighbor sampling. GraphSAGE~\citep{bib4} is a classic method that reduces computational costs by aggregating information through the random sampling of neighbor nodes, facilitating node embedding learning on large-scale graphs. Other methods,  such as GAT~\citep{bib3} and E-ResGAT~\citep{bib43}, enhance this process by introducing importance weights or attention mechanisms. These techniques assign weights based on the degree of neighbor nodes or node similarity, and then aggregate node features using attention weights at each layer. GIN~\citep{bib6} mixes the original features of a graph node with those of its neighbors after each hop aggregation operation, introducing a learnable parameter to adjust its own features, which are then added to the aggregated features of neighboring nodes, ensures that central nodes remain distinguishable from their neighbors. JK-Net~\citep{bib7} and GraphSAGE++~\citep{bib44} strengthen information aggregation in graph neural networks by using node representations from different layers, allowing the network to learn the node's multiscale characteristics more effectively, which allow each node to adaptively select its optimal neighbor aggregation range, not only enhancing the capture of local and global information but also optimizing the common oversmoothing problem in deep networks. However, despite these strategies enhancing node information representation, these models are still limited by their receptive field when dealing with distant neighbor nodes in the graph. This limitation means that capturing information from distant nodes, especially in large-scale or structurally complex graphs, remains a challenge.

Methods based on random walk: DeepWalk~\citep{bib27} is an unsupervised learning technique that combines random walks with language models. It generates node sequences using truncated random walks, similar to sentences. These sequences are then processed through a Skip-gram model to learn vector representations of the nodes, capturing both local and global structural features of the graph. Node2Vec~\citep{bib31}, building on DeepWalk~\citep{bib27}, introduces a flexible biased random walk strategy that balances breadth-first search (BFS) and depth-first search (DFS) strategies, making the exploration of graph structures more flexible. Thus, Node2Vec~\citep{bib31} is able to capture homophily and structural equivalence patterns, improving the representation of community structures and structural roles. InfiniteWalk~\citep{bib32} learns to represent nodes in a graph through a special mathematical transformation of the graph's deep structure. This method, based on extending one of the parameters of the DeepWalk~\citep{bib27} algorithm to infinity, enables it to better capture and utilize information in the graph, performing as well as other advanced methods in tasks such as node classification. Additionally, the MIRW~\citep{bib28} algorithm considers the mutual influences between nodes to better capture network complexity, unlike traditional random walk strategies that treat all nodes and links as equally important, failing to fully reflect the graph's overall structure. CSADW~\citep{bib29} enhances link prediction capabilities in social networks by combining a new transition matrix with structural and attribute similarities. This approach improves traditional techniques, ensuring that random walks are more likely to interact with structurally similar nodes, thereby capturing both structural and non-structural similarities. While these methods are very suitable for capturing community structures and enhancing link prediction capabilities in graphs, they mainly focus on local structural information, have limited perception of global structures, and are less efficient on large-scale graph data.

Methods based on graph transformer: The transformer model has garnered widespread attention due to its outstanding performance in the field of natural language processing, mainly relying on its core component---the self-attention mechanism---to handle dependencies in sequence data. The core idea of the self-attention mechanism is to compute attention scores for each element in the sequence regarding other elements, thereby dynamically adjusting each element's representation. When this mechanism is applied to graph data, it can naturally capture the complex interactions between nodes in the graph. GraphTransformer~\citep{bib24} adopts a unique graphical attention mechanism that uses node adjacency relationships to calculate attention scores, thus making the attention distribution more focused on the local structure of the nodes. This method not only captures direct connections between nodes but also enhances the model's overall perception of graph structures by introducing positional encodings through Laplacian eigenvectors. Graphormer~\citep{bib11} introduced centrality encoding, spatial encoding, and edge encoding, enabling it to better handle graph data structures. TransGNN~\citep{bib8} combines Transformer and graph neural networks, alternating Transformer layers and GNN layers in the model, while introducing three types of positional encodings based on shortest paths, node degrees, and PageRank to help the Transformer layers capture the strusctural information of the graph, enhancing the model's performance in processing graph data. DeepGraph~\citep{bib12} introduced a substructure-based local attention mechanism, incorporating additional substructure tokens in the model and applying local attention to nodes related to these substructures, solving the performance bottleneck faced by traditional graph Transformers in processing deep graph structure data by increasing the number of layers. NodeFormer~\citep{bib10} optimizes the algorithmic complexity of message passing using a kernelized Gumbel-Softmax operator, reducing it from quadratic to linear complexity. NAGphormer~\citep{bib15} and SGFormer~\citep{bib9} are both optimized for processing large-scale graph data, with NAGphormer~\cite{bib15} adopting an innovative Hop2Token module by viewing each node as a sequence composed of its multilevel neighborhood features, and its attention-based readout mechanism optimizes learning of the importance of different neighborhoods, significantly improving the accuracy of node classification. SGFormer~\citep{bib9} simplifies the structure of graph Transformers by adopting a single-layer global attention mechanism to effectively capture latent dependencies between nodes, significantly reducing computational complexity. However, these graph Transformers introduce significant computational and storage overhead to the model through encoding, especially when processing large-scale graph data, which can greatly impact the model's scalability and practicality.

\section{Problem definition}
This section provides some basic definitions.
\begin{definition}
	Given a graph $G = (V,E,X)$, where $V$= \{$v_1$, $v_2$, ..., $ v_N$\} represents a set of $N$ nodes. $N= |V|$ is the total number of nodes in the graph, and $|E|$ is the number of edges. The edge set $E=\{e_{i,j}\}_{i,j=1}^N$ represents all edges in the graph, when $e_{i,j}=1$ if there is an edge between $v_i$  and  $v_j$, and $e_{i,j}=0$ otherwise. The matrix $X \in R^{N\times D}$ denotes the feature matrix for all notdes, with $x_v \in R^D$ representing the feature vector of a node $v$.

\end{definition}

\begin{definition}
	For a given graph with $N$ nodes, the goal of graph representation learning is to develop a mapping function $f : V \rightarrow \mathbb{R}^D$ , that encodes each node in the graph into a low dimensional vector of dimension $D$, where $D \ll |V|$. This mapping ensures that the similarities between nodes in the graph are preserved within the embedding space.
\end{definition}

\begin{definition}

	The feature of vertex \( v \) at the \( i \)th layer is denoted by \( h_v^i \), which is derived from neighborhood aggregation component, and \( h_v^{i}{'} \), which is derived from a \text{Top}-$m$ mechanism aggregation component. The initial feature vectors at layer 0 are \(g_v^0 = h_v^0 = h_v^{0}{'} = X_v \). The output vector \( g_v^i \) for node \( v \) at layer \( i \) is formed by first concatenating \( h_v^i \) and \( h_v^{i}{'} \), and then processing the combined features through a multilayer perceptron (MLP), as expressed by \( g_v^i = MLP ( [h_v^i || h_v^{i}{'}]) \). $\mathcal{N}(v)=\{u \in V:(u,v\in E)\}$ represents the neighborhood set of vertex  $v$ in the graph.
	
\end{definition}

\begin{figure}[h]
	\centering
	\includegraphics[width=\linewidth]{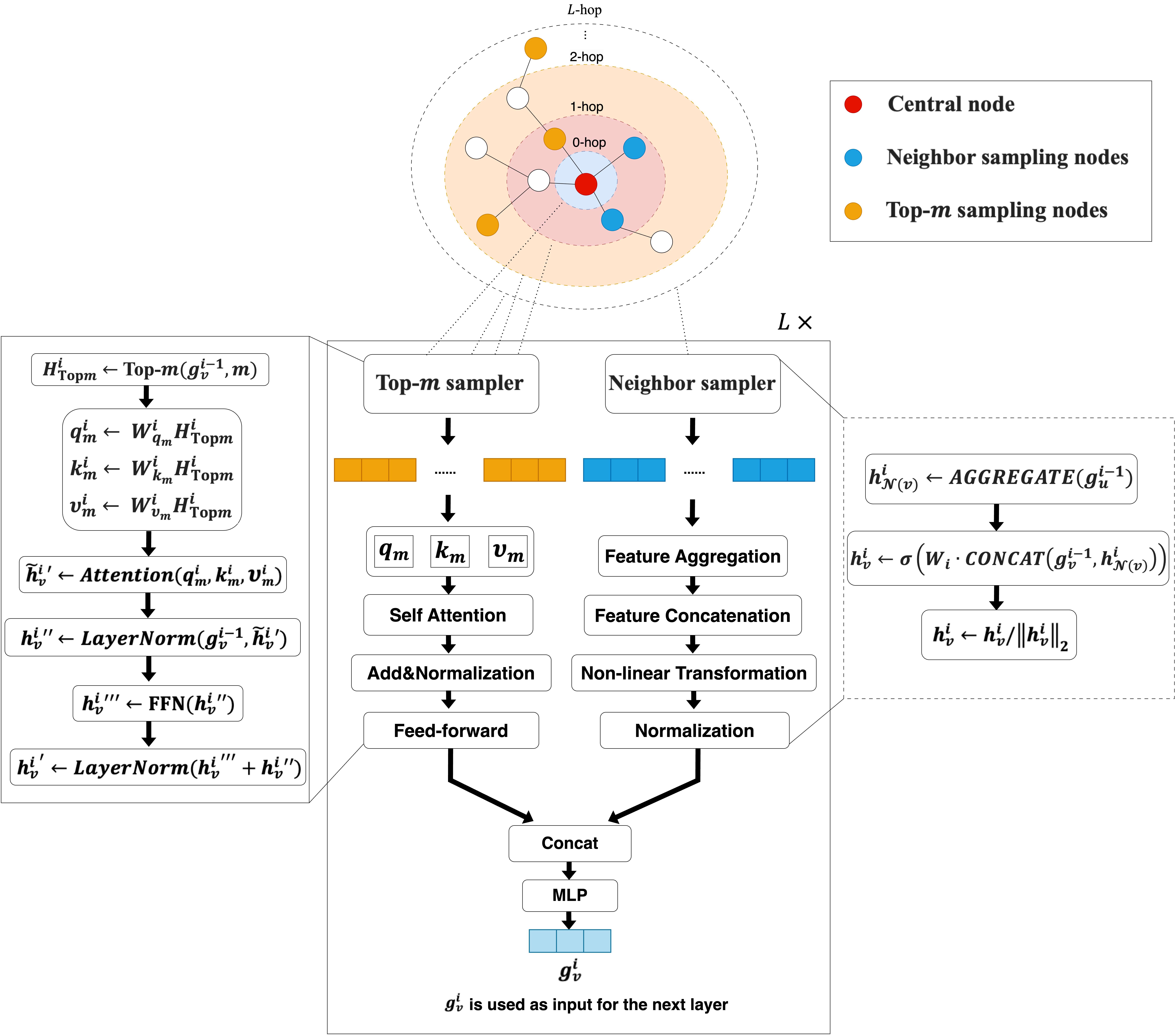}
	\caption{TANGNN Architecture}
	\label{fig1}
\end{figure}
\section{TANGNN}
In this section, we first present the framework of TANGNN, then elaborate on the details of each component of TANGNN, and introduce some variants and algorithm optimizations of this framework.

\begin{algorithm}[htbp]
	\caption{TANGNN Main\_Forward Algorithm}
	\label{algo4}
	\small
	\begin{algorithmic}[1]
		\Require $G(V,E)$; input features$\{X_v,\forall v \in V\}$; depth $K$; Weight matrices $W$; Weight matrices of query, key, value $W_{q_m}$, $W_{k_m}$, $W_{\nu_m}$; Multi-layer perceptron $MLP$
		\Ensure The vector representation $g_v$ for all $\forall v \in V$
		\State	Function Main\_Forward$(X_v, W, W_{q_m}, W_{k_m}, W_{\nu_m}, lr)$
		\State $g_v^0 \leftarrow X_v ,\forall v \in V$;
		\For {$i = 1$ to $L$} 
		\For {$v \in V$} 
		\State $h_v^i \leftarrow$ NeighAggrForward $(g_v^{i-1}, W )$
		\State $h_v^{i}{'} \leftarrow$ TopmAggrForward $(g_v^{i-1}, W_{q_m}, W_{k_m}, W_{\nu_m})$
		\State $g_v^i \leftarrow MLP(Concat(h_v^{i}, h_v^{i}{'} ))$
		\EndFor
		\EndFor
		\State $g_v \leftarrow g_v^L, \forall v \in V$
	\end{algorithmic}
\end{algorithm}
\subsection{Framework of TANGNN}\label{subsec1}

Traditional GNNs only aggregate information from a node's neighbors, thereby limiting their receptive field and affecting their performance. The common practice is to increase the number of layers in GNNs, allowing nodes to aggregate information from farther neighbors, thus expanding the model's receptive field. However, this increase in layers leads to the problem of oversmoothing. Transformer models have a global receptive field, but as the sequence length increases, the computational complexity and storage requirements grow quadratically, so introducing the global attention mechanism of the Transformer model into GNNs faces the same issues. Effectively expanding the receptive field of GNNs and quickly aggregating information is a challenging problem. The key issue is how to effectively select  the appropriate receptive field.

To address the above problems, we have creatively constructed a new GNN framework (as shown in Figure~\ref{fig1}). This GNN has $L$ layers, each consisting of two components: Top-$m$ attention mechanism aggregation and neighbor aggregation.

For the red node $v$ in the graph, based on the node representation calculated from the previous layer, the Top-$m$ nodes most similar to $v$ are computed (the yellow nodes in the graph). Since the node representation aggregates the graph's structure and feature information through the preceding network layer, the Top-$m$ nodes most similar to $v$ are essentially calculated based on the similarity of graph structure and feature information. Therefore, the Top-$m$ attention mechanism component aggregates the graph structure and feature information contained in the Top-$m$ similar nodes to $v$.

The neighbor aggregation component fundamentally performs traditional GNN aggregation. Inspired by GraphSAGE, it uses random sampling of a fixed number of neighbors (the blue nodes in the graph) for aggregation, which can greatly increase the training speed of the model without reducing its performance.

The vector representation $h_{v}^{i}$ generated by the neighbor aggregation in the $i$th layer is concatenated with the vector representation $h_{v}^{i}{'}$ generated by the Top-$m$ attention mechanism aggregation component to obtain the concatenated vector for this layer $[h_{v}^{i}, h_{v}^{i}{'}]$. Through an MLP, $g_{v}^{i}$ is obtained, which is then used as input for the next layer until the $L$th layer to obtain the final vector representation $g_{v}^{L}$, enabling the model to better capture both local and extended neighborhoods'  information.

The Top-$m$ nodes and a fixed number of neighbors of $v$ constitute the receptive field of node $v$. These nodes, based on graph structure and feature information, are most relevant to $v$ and maintain a small scale, addressing the key issue of how to effectively select the appropriate the receptive field. Additionally, the fixed number of receptive field nodes is beneficial for model training and computation.

Algorithm~\ref{algo4} is the forward propagation process of TANGNN. Lines 5 to 6 respectively call the forward propagation functions of the neighborhood aggregation component and the Top-$m$ attention mechanism aggregation component (Algorithms~\ref{algo1} and~\ref{algo2}), to calculate the vector representations $h_v^i$ and $h_v^i{'}$ of node $v$ at the $i$th layer. These are then concatenated and processed through a multilayer perceptron (MLP) to obtain the final vector representation $g_v^{i}$ of node $v$ at the $i$th layer (see line 7 of the code).

\subsection{Model implementation}\label{subsec2}
In this section, we provide a detailed introduction to the implementation of the proposed model, including neighbor aggregation component, Top-$m$ attention mechanism aggregation component, and parameter updates.

\subsubsection{Neighborhood Aggregation}\label{subsubsec1}

\begin{algorithm}[htbp]
	\caption{TANGNN NeighAggrForward Algorithm}
	\label{algo1}
	\small
	\begin{algorithmic}[1]
		\Require $G(V,E)$; input features$\{X_v,\forall v \in V\}$;  Aggregator functions AGGREGATE; Neighborhood Sampling Function $\mathcal{N}$; non-linearity $\sigma $; Weight matrices $W = W_1,...,W_K , \forall i \in \{1,...,K\}$
		\Ensure The vector representation of node $v$ at $i$th is $h_{v} ^ {i}$
		\item 	Function NeighAggrForward$(g_v^{i-1},W)$
		\item \quad $h_{\mathcal{N}(v)}^i$ $\leftarrow$ AGGEREGATE$(g_{u}^{i-1},u \in \mathcal{N}(v) )$
		\item \quad $h_{v}^{i} \leftarrow \sigma (W_i \cdot  CONCAT(g_{v}^{i-1},h_{\mathcal{N}(v)}^i))$
		\item \quad $h_{v}^{i} \leftarrow h_{v}^{i}/||h_{v}^{i}||_2$
		\item Return $h_{v} ^ {i}$
	\end{algorithmic}
\end{algorithm}

Algorithm~\ref{algo1} demonstrates the forward propagation process of neighbor aggregation components. In this process, 
at the $i$th layer, the model firstly gathers a fixed set of neighbors, $\mathcal{N}(v)$, for the node $v$. Then it aggregates the neighbor information $h_{\mathcal{N}(v)}^i \leftarrow AGGEREGATE (h_{u}^{i-1},u \in \mathcal{N}(v) )$ (See line 2 of the code), here, the aggregation function used by the model is the Mean aggregation function. Finally, it updates the vector representation of node $v$ $h_{v}^{i} \leftarrow \sigma (W_i \cdot  CONCAT(h_{v}^{i-1},h_{\mathcal{N}(v)}^i))$ (line 3 of the code), where Sigmod function $\sigma $ is a nonlinear function: $\frac{1}{1+e^{(-x)}} $, it is capable of extracting nonlinear relationships. 

\subsubsection{Top-$m$ Attention Mechanism Aggregation}\label{subsubsec2}

\begin{algorithm}[htbp]
	\caption{TANGNN TopmAggrForward Algorithm}
	\label{algo2}
	\small
	\begin{algorithmic}[1]
		\Require $G(V,E)$; input features$\{X_v,\forall v \in V\}$; Weight matrices of query, key, value $W_{q_m}$, $W_{k_m}$, $W_{\nu_m}$; Feedforward Network $FFN$; Sampling Function Top-$m$; Layer normalization $LayerNorm$
		\Ensure The vector representation of node $v$ at $i$th is $h_{v} ^ {i}{'}$ 
		\item 	Function TopmAggrForward$(g_v^{i-1}, W_{q_m}, W_{k_m}, W_{\nu_m})$
		\item \quad $H_{\text{Top-}m}^{i}$ $\leftarrow$ Top-$m$$({g_{v}^{i-1}}, m )$
		\item \quad According to Eq.~(\ref{equation1}) to~(\ref{equation3}), obtain the values of $q_m^i$, $k_m^{i}$, and $\nu_m^{i}$
		\item \quad $\tilde{h}_v^{i}{'}$ $\leftarrow$ $Attention(q_m^{i}, k_m^{i}, \nu_m^{i})$
		\item \quad $h_v^{i}{''}$ $\leftarrow$ $LayerNorm(g_v^{i-1}, \tilde{h}_v^{i}{'})$
		\item \quad $h_v^{i}{'''}$ $\leftarrow$ $FFN(h_v^{i}{''})$
		\item \quad $h_v^{i}{'}$ $\leftarrow$ $LayerNorm(h_v^{i}{'''} + h_v^{i}{''})$
		\item Return $h_{v} ^ {i}{'}$
	\end{algorithmic}
\end{algorithm}

Algorithm~\ref{algo2} demonstrates the forward propagation process of the Top-$m$ attention mechanism aggregation component. During this process, apply the Top-$m$ strategy ( (Details are shown in section~\ref{subsec3}) ) to the output of the previous layer of each node $v$, which is $g_v^{i-1}{'}$, to select the $m$ nodes with the highest similarity and obtain the representation $H_{\text{Top-}m}^{i}$ (see line 5 of the code). As shown in Eq.~(\ref{equation1}) to~(\ref{equation3}), the self-attention mechanism first maps it to the query ($q$), key ($k$), and value ($\nu$) vectors:

\begin{equation}\label{equation1}
	q_m^{i} = W_{q_m}^{i}H_{\text{Top-}m}^{i}
\end{equation}

\begin{equation}\label{equation2}
	k_m^{i} = W_{k_m}^{i}H_{\text{Top-}m}^{i}
\end{equation}

\begin{equation}\label{equation3}
	\nu_m^{i} = W_{\nu_m}^{i}H_{\text{Top-}m}^{i}
\end{equation}

Here, $W_{q_m}^{i}, W_{k_m}^{i}, W_{\nu_m}^{i}$ are the learnable weight matrices for the $i$th layer. The query, key, and value matrices corresponding to each head typically have smaller dimensions, denoted as $d_q, d_k, d_\nu$ espectively. Next, compute the dot product of the query vector with the key vectors, scaled by a factor of $\sqrt{d_k}$, and then apply softmax to obtain the attention weights, as shown in Eq.~(\ref{equation4}):

\begin{equation}\label{equation4}
	A_{vu}^{i} = \frac{\exp \left(q_{mv}^{i} \cdot (k_{mu}^{i})^T / \sqrt{d_k}\right)}{\sum_{u=1}^m \exp \left(q_{mv}^{i} \cdot (k_{mu}^{i})^T / \sqrt{d_k}\right)}
\end{equation}

Here, $A_{vu}^{i}$ represents the attention weight from node $v$ to node $u$ at the $i$th layer, where $m$ is the number of nodes obtained from the Top-$m$ attention mechanism component. Then we use Eq.~(\ref{equation5}) to calculate the output of the head:

\begin{equation}\label{equation5}
	Head_{v}^{i} = \sum_{u=1}^{m} A_{vu}^{i} \nu_{mu}^{i}
\end{equation}

The output vectors linearly transformed by another weight matrix $W^{o}$, as shown in Eq.~(\ref{equation6}).

\begin{equation}\label{equation6}
	\tilde{h}_v^{i}{'} = Head_{v}^{i}W^o
\end{equation}

Subsequently, after layer normalization and a feed-forward neural network, the output is merged $h_v^{i}{''}$ $\leftarrow$ $LayerNorm$ $(g_v^{i-1}, \tilde{h}_v^{i}{'})$ (line 5 of the code), and then it is fed into a feed-forward neural network (line 6 of the code). Finally, the output of the current layer is completed through a second layer normalization (line 7 of the code), obtaining the vector representation $h_v^{i}{'}$ of node $v$ at the $i$th layer.

\begin{algorithm}[htbp]
	\caption{TANGNN Backpropagation Algorithm}
	\label{algo5}
	\small 
	\begin{algorithmic}[1]
		\Require $G(V,E)$; input features $\{X_v, \forall v \in V\}$; depth $L$; learning rate $lr$; Node label $y$
		\Ensure The Weight parameters after updating Params [$W,$ $W_{q_m},$ $W_{k_m}, W_{\nu_m}$]
		\State Input $G(V,E)$
		\State Initialize the weight parameter $W, W_{q_m}, W_{k_m}, W_{\nu_m}$ with Xavier initialization
		\Repeat
		\State $g_v \leftarrow \text{Main\_Forward}(X_v, W, W_{q_m}, W_{k_m}, W_{\nu_m}, lr)$
		
		\State $p \leftarrow softmax(g_v)$
		\State  $q \leftarrow one\_hot(y)$
		\State Calculate the $loss$ value using equation (\ref{equation7}), $loss \leftarrow L(p,q)$
		\State  gradient $\leftarrow gradients(loss,W)$
		\State $W \leftarrow lr$ $*$ gradient 
		\Until{model has completed convergence}
	\end{algorithmic}
\end{algorithm}

Algorithm~\ref{algo5} shows the backpropagation process of TANGNN. It firstly calls Algorithm~\ref{algo4} to compute the node vector representation $g_v$ (see line 4 of the code). Lines 5 to 7 involve the calculation of the loss value, starting with a softmax operation on each node's vector representation to convert it into a probability distribution $p$, then converting the node's label $y$ into one\_hot encoding, and finally computing the loss value through the loss function. Lines 8 to 9 involve calculating gradients to implement the update of weight parameters.

$W$ is the training parameter for the neighbor aggregation component of Algorithm 1, and $W_{q_m}, W_{k_m}, W_{\nu_m}$ are the weight parameters for the Top-$m$ attention mechanism aggregation component, all of which are obtained through training with the following objective function:
\begin{equation}\label{equation7}
	L(p,q) = -\sum (p(v)logq(v) + (1-p(v))log(1-q(v)))
\end{equation}

Where the probability distribution $p$ is the desired output, and the probability distribution $q$ is the actual output. This objective function primarily measures the distance between the actual output (probability) and the desired output (probability). A smaller value indicates that the two probability distributions are closer to each other.

\begin{figure}[htbp]
	\centering
	\includegraphics[width=0.8\textwidth]{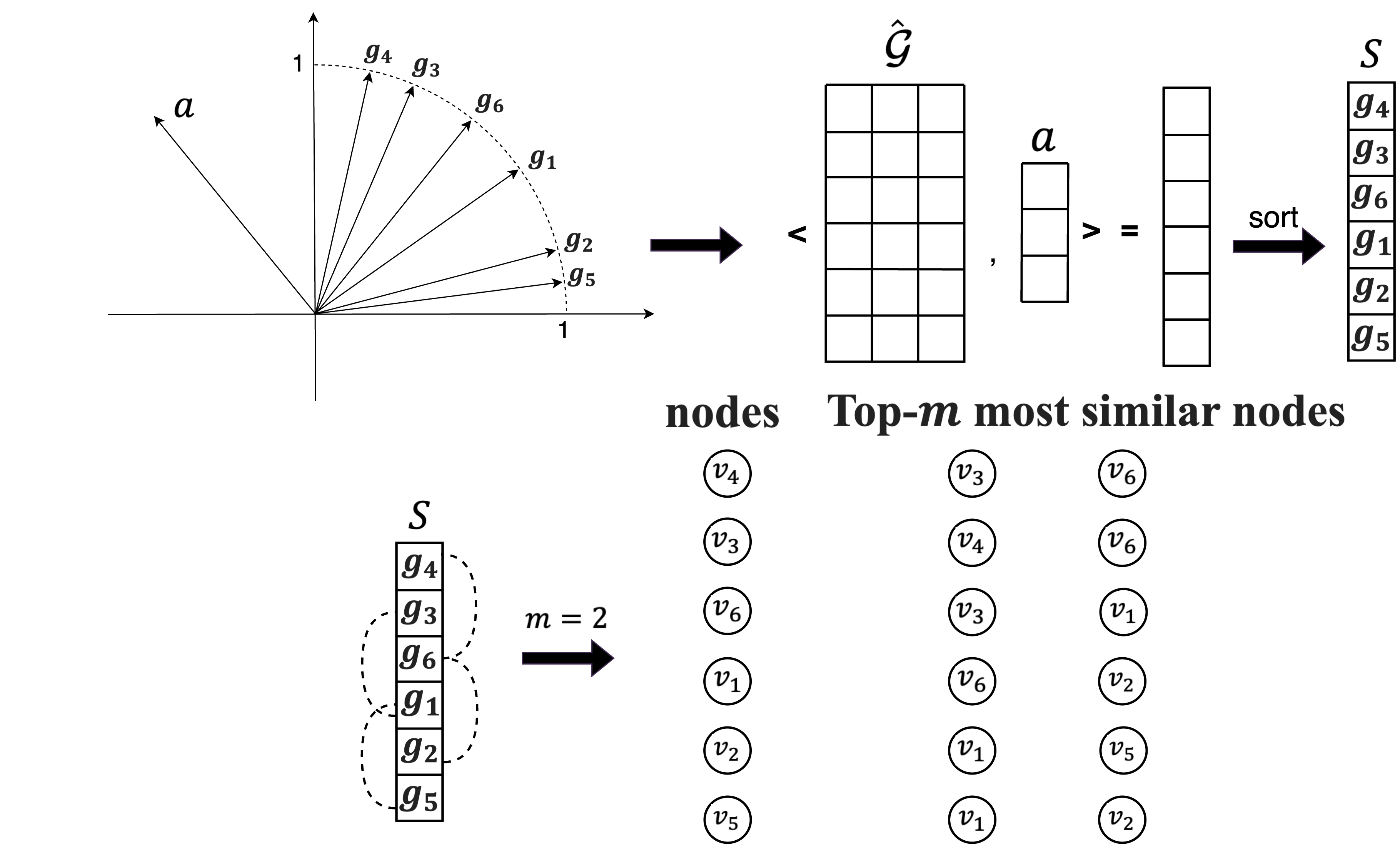}
	\caption{Illustration of the Top-$m$ Efficient Algorithm. An auxiliary vector $a$ is introduced into the model to calculate the cosine similarity between node vector representations and the auxiliary vector, determining the similarity between nodes. This method utilizes the transfer property of similarity, reducing the necessary computational complexity. The value of $m$ is set to 2, the two most similar nodes to each node are its two nearest neighbors. The boundary nodes ($g_4$ and $g_5$) have their two nearest nodes as their most similar nodes.}
	\label{fig2}
\end{figure}

\subsection{Top-\texorpdfstring{$\bm{m}$}{m} Efficient Algorithm}\label{subsec3}
Calculating the similarity between all node pairs is a computationally expensive task, especially in large graph data. This method not only has high computational complexity, with a time complexity of $O(N^2)$, but also significantly increases the demands on memory and storage. Inspired by DiglacianGCN~\citep{bib14}, we introduce an auxiliary vector $a$ into the model, which indirectly estimates the similarity between nodes by comparing it with the node features' similarity. This approach utilizes the transitive property of similarity, i.e., the higher the similarity between two nodes and $a$, the more similar these two nodes are. Specifically, to further optimize efficiency and enhance the model's robustness, The Top-$m$ algorithm in this paper is specifically conducted on positive vectors, avoiding interference from potential noise and irrelevant information. The node vector $g_n$ is first normalized using L2 normalization, represented as $\hat{g}_n= \text{LN}_2(g_n) = \frac{g_n}{\|g_n\|_2}$. Let $ \hat{\mathcal{G} } = \{\hat{g}_n\}_{n=1}^N, \bar{g} = \frac{1}{N} \sum_{n=1}^N \hat{g}_n$. In this way, we ensure that the norm of the node vector is 1. We will randomly initialize $a$ as a $d$-dimensional vector, where ${a \neq \bar{g}}$ and ${a \neq 0}$. The update rule for $a$ is shown in Eq. (\ref{equation8}) to ensure that $a$ is not aligned with the feature vector of any specific node.

\begin{equation}\label{equation8}
	a = \text{LN}_2\left(a - (a^T\bar{g})\bar{g}\right)
\end{equation}

Clearly, $a$ is orthogonal to the unit vector $\bar{g}$. Therefore, feature vectors with high similarity will be mapped close to the position of $a$. The proof of $a$ being orthogonal to $\bar{g}$ is as follows:

\begin{equation}\label{equation9}
	(a - (a^T \bar{g}) \bar{g})\cdot\bar{g} = a \cdot \bar{g} - (a^T \bar{g}) \bar{g} \cdot \bar{g} = a \cdot \bar{g} - (a \cdot \bar{g})\|\bar{g}^2\| = a \cdot \bar{g} - a \cdot \bar{g} = 0
\end{equation}

As shown in Figure \ref{fig2}, we use the auxiliary vector $a$ to calculate node similarity. For each sampled node vector $g_n$, we use Eq.~(\ref{10}) to compute its similarity score with the auxiliary vector $a$ using cosine similarity. Here, $s_n$ is the similarity score between  $g_n$ and $a$, $a^T$ represents the transpose of $a$. 

\begin{equation}\label{10}
	s_n = similarity(a, g_n) = a^T\hat{g}_n
\end{equation}

The calculated similarity scores $s_n$ are used to sort the node indices, producing a sorted list $S$ as shown in Eq.~(\ref{11}), where $N$ is the total number of sampled nodes. The nodes are sorted based on their similarity scores in descending order, resulting in a list of nodes ranked from most to least similar. As illustrated in Figure \ref{fig2}, the sorted list $S$ represents each node connecting to its Top-$m$ most similar nodes, where the value of $m$ is set to 2. Thus, the two most similar nodes to each node are its two nearest neighbors. The boundary nodes (in the graph, $g_4$ and $g_5$)have their two nearest nodes as their most similar nodes.

\begin{equation}\label{11}
	S = \text{sort}(\{s_n\}_{n=1}^N)
\end{equation}

By introducing the auxiliary vector $a$, and utilizing the transitive property of similarity, the model no longer needs to directly compute the similarity between all pairs of nodes. Instead, it only calculates the similarity of each node with $a$, reducing the time complexity from $O(N^2)$ to $O(N)$. This method enables the model to be more effectively applied to large-scale graph data.

\begin{figure}[htbp]
	\centering
	\includegraphics[width=0.7\textwidth]{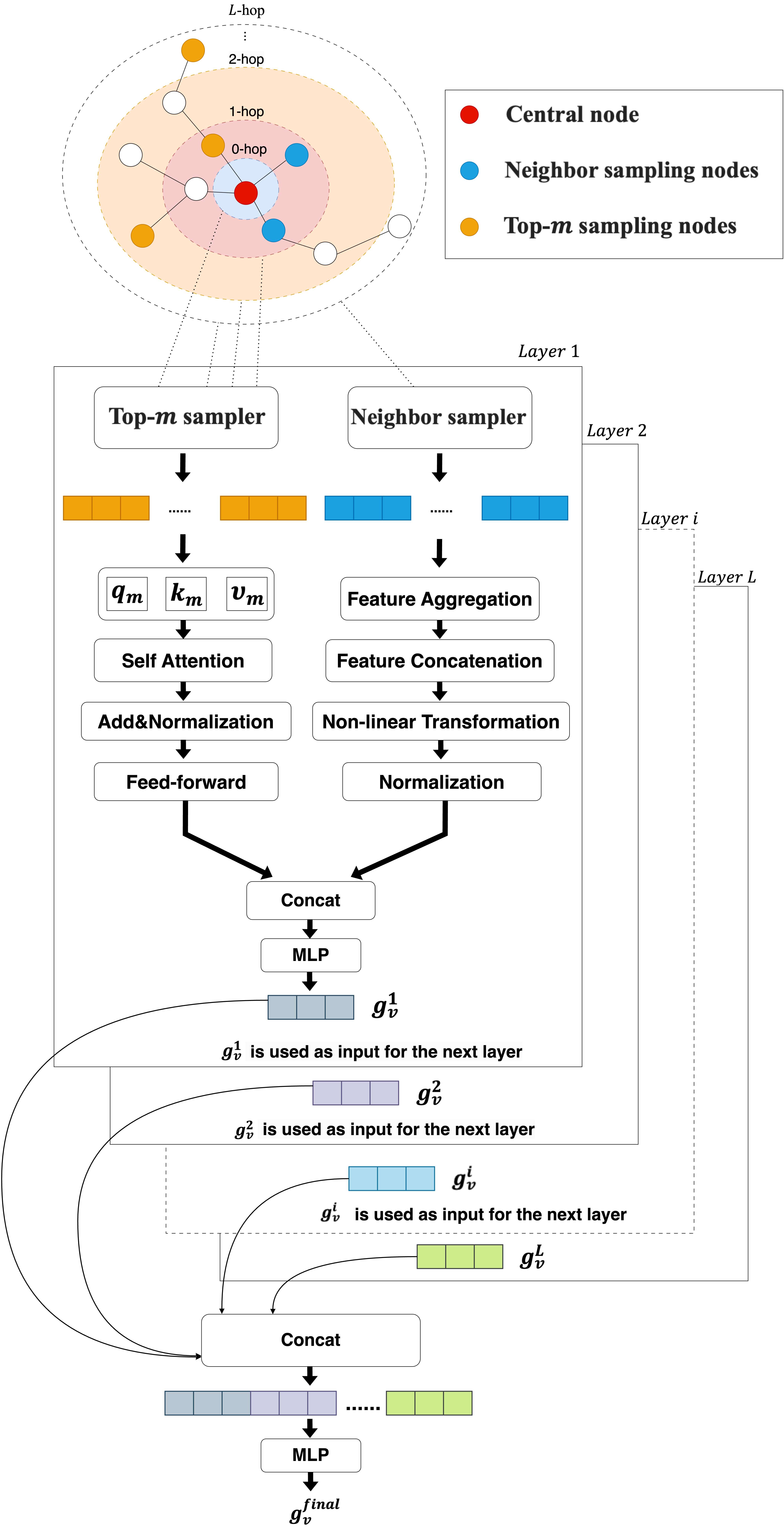}
	\caption{TANGNN-LC  architecture}
	\label{fig3}
\end{figure}

\subsection{Model Optimization TANGNN-LC}\label{subsec4}
In this section, we present several enhancements to our model. 
Building upon the TANGNN, we have introduced a layer concatenation strategy. As shown in Figure~\ref{fig3}, in this model, the outputs from different layers are not only passed to the next layer but are also concatenated in the final stage of the model. The concatenated vector [$g_v^1, g_v^2,..., g_v^L$] is then processed through a multi-layer perceptron, resulting in the final vector representation $g_v^{final}$ of vertex $v$. This design allows the model to consider information from all layers in its final output, capturing deep structural features of the graph while also retaining key local information, thus enhancing the model's performance on complex graph data.

\subsection{Model Variants}\label{subsec5}
In this section, we have defined different methods for updating nodes, including TANGNN-NA (Neighborhood Aggregation As Input), TANGNN-TA (Top-$m$ Attention As Input), and TANGNN-FLC (Final Layer Concatenation). Through experimental comparison, it was found that these algorithms also have certain competitiveness compared to baseline algorithms.

\begin{figure}[htbp]
	\centering
	\includegraphics[width=0.8\textwidth]{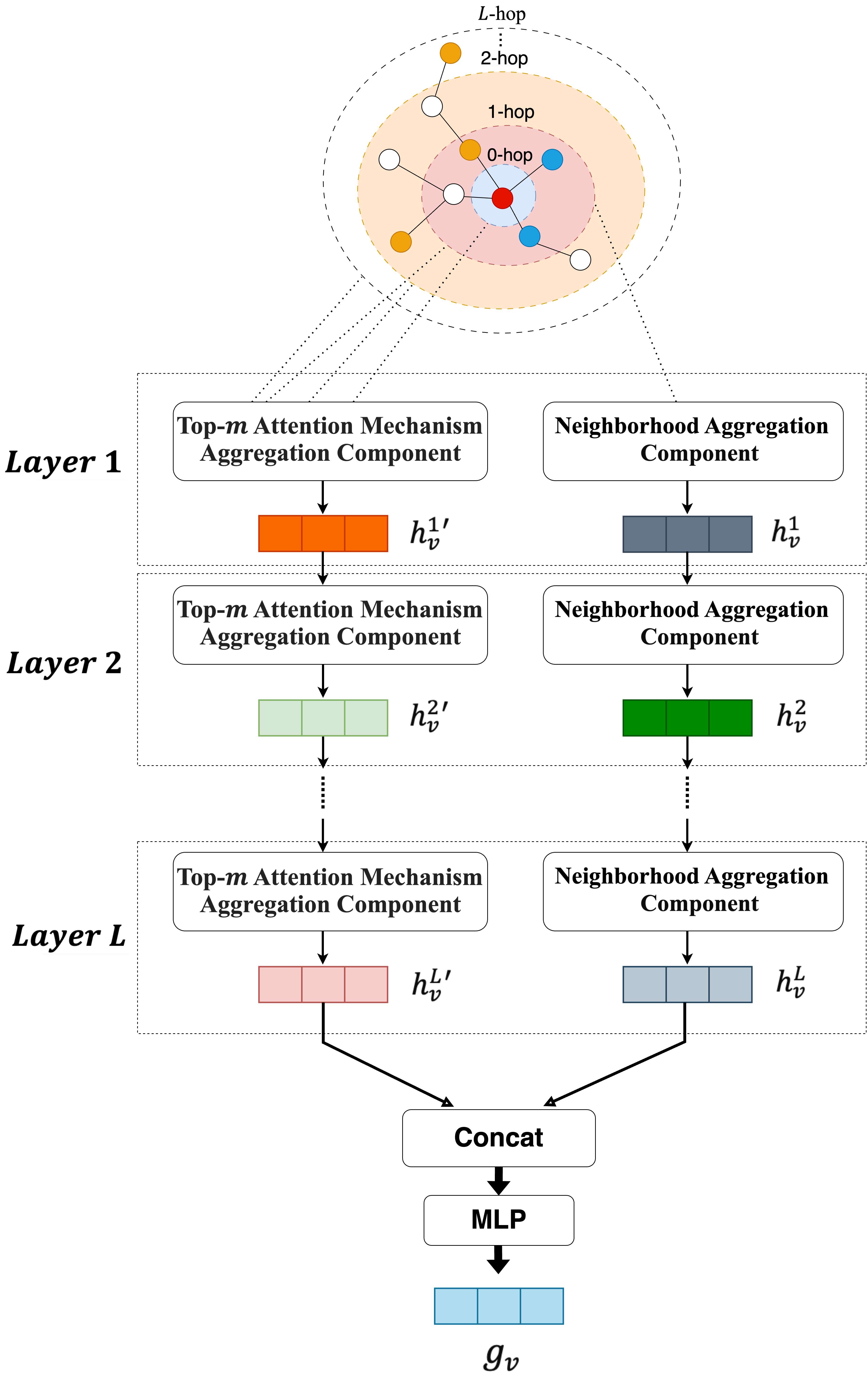}
	\caption{TANGNN-FLC  architecture}
	\label{fig4}
\end{figure}
\subsubsection{TANGNN-FLC(Final Layer Concatenation)}\label{subsubsec4}

As shown in Figure \ref{fig4}, unlike TANGNN, TANGNN-FLC concatenates the node information processed by the two components only in the final layer. The neighborhood aggregation component produces the feature vector representation $h_v^L$  of the target node at the $L$th layer, and the Top-$m$ attention mechanism aggregation component then computes based on the output  $h_v^{L-1}{'}$ from the previous layer's , produces the feature vector representation $h_v^L{'}$ of the target node at the $L$th layer. These two feature vectors are concatenated and then processed through a multilayer perceptron to form the final vector representation $g_v$ of the target node.

\subsubsection{TANGNN-NAI(Neighborhood Aggregation As Input)}\label{subsubsec5}
TANGNN-FLC cannot effectively retain the information of each layer, and as the number of layers increases, the problem of oversmoothing may occur. To address this, we designed TANGNN-NAI based on the TANGNN framework. $h_v^i$ is the vector representation of the neighborhood aggregation component at the $i$th layer, and $h_v^i{'}$ is the vector representation of the Top-$m$ attention mechanism aggregation component at the $i$th layer. These two vector representations are concatenated to form the vector representation [$h_v^i$, $h_v^i{'}$], which serves as the input for the next layer of the neighborhood aggregation component, while $h_v^i{'}$ serves as the input for the next layer of the Top-$m$ attention mechanism aggregation component, continuing until the final layer and then passing through a multilayer perceptron to obtain the final vector representation $g_v^L$. This method effectively captures the structural information of the graph.

\subsubsection{TANGNN-TAI (Top-$m$ Attention As Input)}\label{subsubsec6} 
The only difference between TANGNN-TAI and TANGNN-NAI is that after obtaining the concatenated vector representation at the $i$th layer, different vector representations are input into different components. The concatenated vector representation serves as the input for the next layer of the Top-$m$ attention mechanism aggregation component, while the vector representation $h_v^i$ from the $i$th layer of the neighborhood aggregation component serves as the input for the next layer of the neighborhood aggregation component.

\section{Experiments}
We chose the PyTorch Geometric (PyG) framework for the implementation of our models, renowned for its suitability for processing graph-structured data through deep learning techniques. PyG excels in managing large-scale graph datasets and offers a comprehensive suite of optimized layers for graph neural networks. We carried out all our experiments on a system equipped with an Apple M1 Pro (14 Core) @ 3.20Ghz processor, ensuring ample computational power for the demanding tasks of training and testing our models.

\subsection{Experimental settings}\label{subsec6}

Parameter settings: For all algorithms, the size of the training set is set from 0.1 to 0.9, with the remainder serving as the test set. The learning rate (lr) of the model is set to 0.001. The batch size is set at 128. The number of layers $L$ for our model is set to 2 due to the increased computational burden associated with higher numbers of layers. The value of $m$ in the Top-$m$ attention mechanism aggregation component is set to 30. In the neighbor aggregation component, the aggregation method uses the Mean aggregation function, and the sampling number of the two layers is set to [20,10] (20 for the first layer and 10 for the second layer). To update model parameters, we employ the backpropagation algorithm in conjunction with gradient descent. In each training iteration, the gradients of the model parameters are calculated using the chain rule, and these gradients are used to update the parameters towards minimizing the loss function. Specifically, we utilize the Adam optimizer, which integrates adaptive learning rates with momentum to speed up convergence and improve generalization. Additionally, to boost the model’s representational power and ability to generalize, we apply regularization techniques. Specifically, we use L2 regularization to limit the size of the model parameters, which helps prevent overfitting. By penalizing the L2 norm of the parameters, we encourage the model to select simpler solutions, thereby improving the model's generalization performance.

\subsubsection{DATASETS}\label{subsubsec7}
We conducted evaluations using eight well-established datasets frequently employed for vertex embeddings. The details of these datasets are shown in Table~\ref{tab1} and~\ref{tab2}, including a variety of types such as social networks, protein datasets and citation networks. Both $Cora$~\citep{bib17} and $Pubmed$~\citep{bib17} are recognized as a benchmark in citation network analysis, where each vertex is represented by a bag-of-words from the paper and labeled by its academic subject. $Amazon$~\citep{bib17} models the interactions between users and products, representing users and products as nodes and their transactions as graph edges. $Citeseer$~\citep{bib17} represents a citation network where each node holds specific characteristic information pertaining to the literature. $Reddit$~\citep{bib4} is extensively utilized for research in social network analysis and natural language processing fields. It contains post data from the Reddit platform, organized by community (Subreddit). In this dataset, each node represents a post, and the edges in the graph represent direct links between posts or explicit interactions between users,  its labels correspond to the community to which the post belongs. The $ZINC$~\citep{bib25} is a large chemical database containing millions of drug compounds' 3D structures and attribute information. In graph neural network research, ZINC is commonly used to test molecular property prediction models, such as solubility, toxicity, or biological activity of molecules. The $QM9$~\citep{bib26} includes nearly 134k stable small molecules' quantum chemical properties, with each molecule composed of no more than 9 atoms (C, H, O, N, F). It provides computed chemical properties of molecules, such as molecular orbital energies, dipole moments, free energies, etc. The $ArXivNet$ is derived from unarXive~\citep{bib40}, where papers are extracted to construct its own citation network. The nodes in this dataset represent ArXiv academic papers, directed edges represent citations between documents. We use the pre-trained model SPECTER~\citep{bib50} to extract semantic representations as node features based on the titles and abstracts of papers. And we utilize the DictSentiBERT~\citep{bib49} model to annotate the sentiment polarity (positive, neutral, negative) of citations. The relevant processing code and data have been published on the aforementioned Github website.

\begin{table}[htbp]
	\centering
	\caption{Dataset statistics for node classification, link prediction, and sentiment analysis tasks}\label{tab1}%
	\begin{tabular}{cccc}
		\toprule
		Dateset & Vertices & Edges & Labels \\
		\midrule
		Cora &  2707 & 5429 & 7 \\
		Citeseer & 3327 & 4732 & 6 \\
		PubMed & 19717 & 44327 & 3 \\
		Amazon & 13381 & 245778 & 10 \\
		Reddit & 232965 & 114615892 & 41 \\
		ArXivNet(ours) & 142706 & 371231 & 3 \\
		\botrule
	\end{tabular}
\end{table}

\begin{table}[htbp]
	\centering
	\caption{Dataset statistics for graph regression tasks}\label{tab2}%
	\begin{tabular}{cccc}
		\toprule
		Dateset & Graphs & Avg Nodes & Avg Edges \\
		\midrule
		ZINC &  12000 & 23.2 & 49.8 \\
		Citeseer & 130831 & 18.03 & 37.3 \\
		\botrule
	\end{tabular}
\end{table}

\subsubsection{BASELINE ALGORITHM}\label{subsubsec8}
We use the following representative algorithms in comparative experiments. For all models, the learning rate is set to 0.001 and the batch size is set to 128. The number of layers (hops) for the GNN models GCN, GraphSAGE, SGCN, GAT, JK-Net, GIN, TransGNN, SAGEFormer, and NAGphormer is uniformly set to 2.

$GCN$~\citep{bib1} is one of the initial graph convolutional network models proposed, utilizes adjacency and node feature matrices to update and propagate node features. At each layer, the feature representation of a node is updated by aggregating features from its neighboring nodes.

$SGCN$~\citep{bib5} represents a streamlined version of the graph convolutional neural network. It simplifies the graph convolution process to a single linear operation by eliminating nonlinear transformations and polynomial fitting. This reduction in complexity and computational demand enhances the model’s interpretability and efficiency.

$GraphSAGE$~\citep{bib4} is a graph convolutional neural network that uses an aggregation function to collect neighbor information around vertices. Through training, the algorithm updates this information, allowing vertices to access higher-order neighbor information as the number of iterations increases.

$GAT$~\citep{bib3} is a neural network model based on graph attention mechanisms. Its core concept revolves around learning the relationships between nodes and the importance of features between them, utilizing attention mechanisms to dynamically calculate the influence weights of each node on its neighboring nodes.

$JK$-$Net$~\citep{bib7} is a framework designed to enhance the performance of graph neural networks (GNNs) on tasks like node classification and graph classification. The fundamental concept of it involves leveraging jumping connections to amalgamate information from various levels of graph convolutional layers. This integration helps the model to more effectively learn and represent graph data.

$GIN$~\citep{bib6}: In the task of graph isomorphism detection, the objective is to ascertain if two graphs are isomorphic, meaning there exists a one-to-one correspondence between their nodes that preserves the connectivity of edges. GIN learns graph representations by integrating the features of nodes with those of their neighbors and applying a multilayer perceptron to introduce non-linear transformations to these combined features.

$Graphormer$~\citep{bib11} utilizes the standard Transformer architecture, introducing key structural encodings such as centrality encoding, spatial encoding, and edge encoding to optimize the performance of graph representation learning. This enhances its ability to process graph structure data and improves its expressiveness.

$TransGNN$~\citep{bib8} employs an attention sampling strategy based on semantic and structural information, selecting nodes most relevant to the central node for information aggregation. It alternates between Transformer layers and GNN layers within the model, using Transformer layers to expand the receptive field of GNNs, and GNN layers to aid the Transformer in better understanding the graph structure.

$NAGphormer$~\citep{bib15} introduces a neighborhood aggregation module that can aggregate features from multiple hops and converts each hop's features into a token. Each token represents the central node's neighborhood information at different hop distances, enabling the Transformer to be effectively applied to node classification tasks in large-scale data.

$SAT$~\citep{bib13} enhances the effectiveness of graph node representations by incorporating structural information into the self-attention mechanism of the Transformer. Compared to traditional methods that focus only on node features, SAT uses extracted local subgraph structure information for attention computation, enhancing the structural awareness of node representations.

$DeepGraph$~\citep{bib12} incorporates additional substructure tokens into the model and uses local attention to deal with nodes related to specific substructures, addressing the performance bottlenecks faced by graph transformers when increasing the number of layers.

$SGFormer$~\citep{bib9} uses a simple single layer attention mechanism, which, by eliminating the need for positional encoding and complex graph preprocessing steps, simplifies the model architecture. This makes information propagation more efficient and reduces computational overhead.

\subsection{Experimental result}\label{subsec9}
In this section, we assess the effectiveness of our methods across tasks such as link prediction, classification, and visualization. We also conduct comparisons with other baseline algorithms to illustrate the advantages of our approaches.
\subsubsection{CLASSIFICATION TASK}\label{subsubsec9}
The objective of vertex classification tasks is to accurately predict the labels for each node in the network.
\begin{figure}[htbp]
	\centering
	\includegraphics[width=\linewidth]{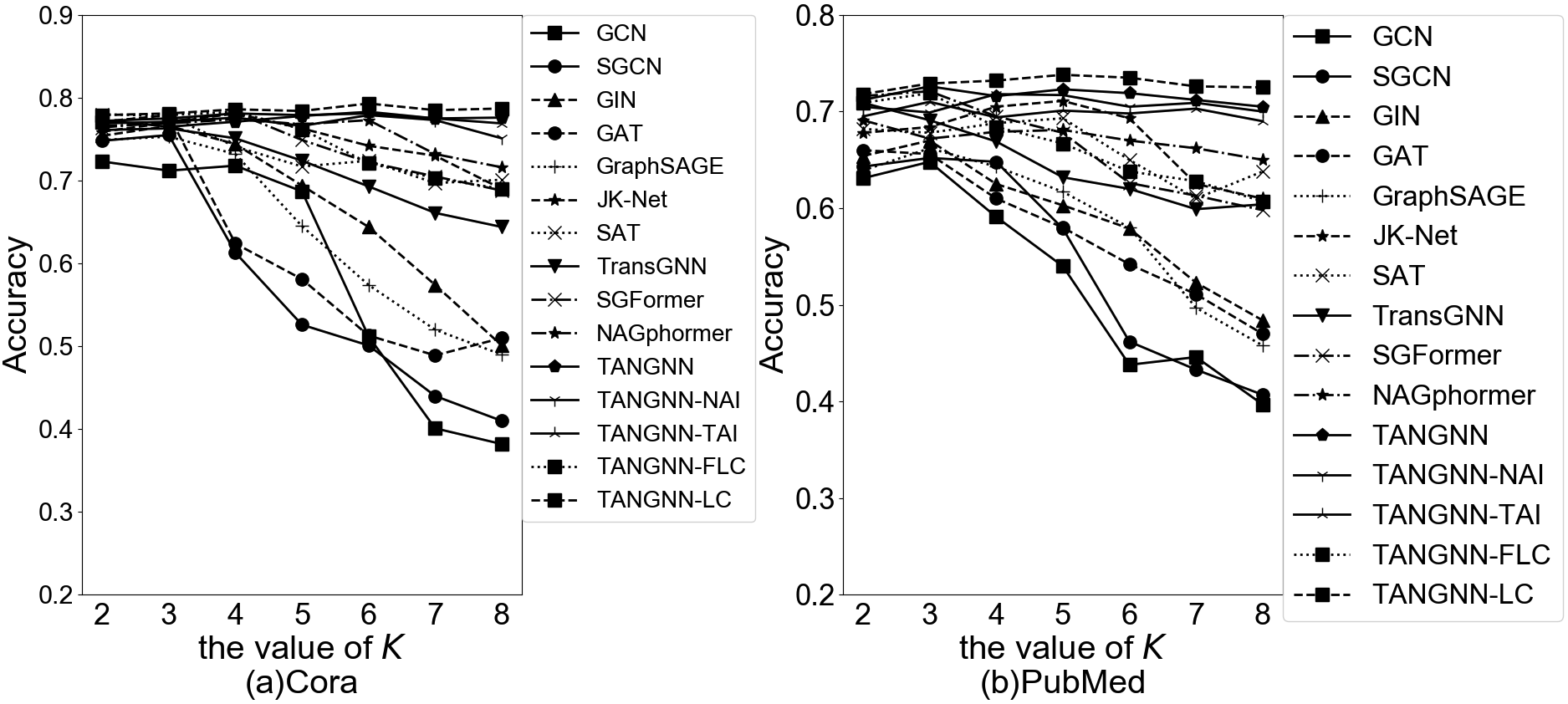}
	\caption{Accuracy values for vertex classification tasks on Cora and PubMed datasets vary with $K$ values}
	\label{fig5}
\end{figure}\textbf{}

Table~\ref{tab3} to~\ref{tab5} shows our method and several other algorithms performing classification tasks on three real datasets. The experimental results show that TANGNN-LC achieved good results on all datasets. On the Cora dataset, TANGNN-LC improved performance by 12\% compared to GraphSAGE (when the training dataset is 50\%). On the PubMed dataset, TANGNN-LC improved by 2\% compared to TANGNN (when the training dataset is 10\%). Overall, compared to other algorithms, TANGNN-LC and TANGNN both have achieved better performance in classification tasks.

Figure~\ref{fig5} illustrates how varying the parameter $K$ affects the accuracy of different algorithms when they are tasked with classification on the Cora and PubMed datasets. The findings indicate that TANGNN maintains stability without showing a significant decline in performance with increasing $K$ values. We can see that TANGNN-LC have achieved the highest accuracy on the pubmed dataset when the value of $K$ is 5. As the depth increases, TransGNN, SGFormer, and NAGphormer show a downward trend due to the influence of GNN entities. JK-Net combines information from various levels of graph convolutional layers using skip connections, thereby retaining more node information. Therefore, the effect of JK-Net on the performance is not significant when the value of $K$ increases. GCN, SGCN and other GCN variants show a significant downward trend in Acc values after the value of $K$ reaches 4.

\begin{table}[h]
	\centering
	\setlength{\tabcolsep}{1.9pt} 
	\caption{The Auroc value obtained by completing vertex classification tasks on the Cora dataset. The best result is marked in bold.}\label{tab3}%
	\footnotesize 
	\begin{tabularx}{\linewidth}{cccccccccc}\toprule
Algorithm & 10\% & 20\% & 30\% & 40\% & 50\% & 60\%& 70\%& 80\% & 90\%\\ \hline
GCN &  0.8901 & 0.8932 & 0.8944 & 0.8970 & 0.9011 &0.9025 & 0.9017 & 0.8952 & 0.8978 \\
GraphSAGE & 0.9013 & 0.9025  & 0.9073 & 0.9084 & 0.9056&0.9120 & 0.9098 & 0.9105 & 0.9100 \\
SGCN & 0.8992 & 0.8989  & 0.8974 & 0.9027 & 0.9038&0.9061 & 0.9023 & 0.9090 & 0.9009 \\
GAT & 0.8997 & 0.9003  & 0.9065 & 0.9052 & 0.9114&0.9152 & 0.9061 & 0.9107 & 0.9105 \\
GIN & 0.9068 & 0.9123  & 0.9144 & 0.9137 & 0.9250&0.9243 & 0.9284 & 0.9209 & 0.9195 \\
JK-Net & 0.9235 & 0.9262  & 0.9310 & 0.9313 & 0.9289&0.9354 & 0.9332 & 0.9268 & 0.9279 \\
Graphormer & 0.9245 & 0.9279  & 0.9384 & 0.9369 & 0.9366&0.9402 & 0.9387 & 0.9393 & 0.9400 \\
SAT & 0.9312 & 0.9383  & 0.9352 & 0.9418 & 0.9443&0.9385 & 0.9422 & 0.9454 & 0.9437\\
TransGNN & 0.9461 & 0.9454  & 0.9470 & 0.9543 & 0.9552&0.9568 & 0.9537 & 0.9526 & 0.9549\\
SGFormer & 0.9570 & 0.9542  & 0.9611 & 0.9626 & 0.9633&0.9689 & 0.9703 & 0.9724 & 0.9711\\
DeepGraph & 0.9487 & 0.9525  & 0.9546 & 0.9574 & 0.9588&0.9633 & 0.9671 & 0.9687 & 0.9690\\
NAGphormer & 0.9473 & 0.9536  & 0.9612 & 0.9597 & 0.9604&0.9638 & 0.9645 & 0.9610 & 0.9588\\
TANGNN-FLC & 0.9439 & 0.9480  & 0.9544 & 0.9523 & 0.9556&0.9539 & 0.9611 & 0.9574 & 0.9586\\
TANGNN-NAI & 0.9428 & 0.9492  & 0.9568 & 0.9573 & 0.9590&0.9600 & 0.9624 & 0.9612 & 0.9624\\
TANGNN-TAI & 0.9403 & 0.9469  & 0.9537 & 0.9585 & 0.9602&0.9633 & 0.9654 & 0.9668 & 0.9631\\
TANGNN & 0.9451 & 0.9493  & 0.9662 & 0.9623 & 0.9679&0.9686& 0.9657 & 0.9630 & 0.9645\\
		TANGNN-LC & \textbf{0.9657} & \textbf{0.9628}  & \textbf{0.9692} & \textbf{0.9703}& \textbf{0.9735}&\textbf{0.9756} & \textbf{0.9745} & \textbf{0.9737} & \textbf{0.9710}\\
		\bottomrule
	\end{tabularx}
\end{table}

\begin{table}[h]
	\centering
	\setlength{\tabcolsep}{1.9pt} 
	\caption{The Auroc value obtained by completing vertex classification tasks on the Citeseer dataset. The best result is marked in bold.}\label{tab4}%
\footnotesize 
	\begin{tabularx}{\linewidth}{cccccccccc}\toprule
		Algorithm & 10\% & 20\% & 30\% & 40\% & 50\% & 60\%& 70\%& 80\% & 90\%\\ \hline
		GCN &  0.6971 & 0.7002 & 0.7013 & 0.7026 & 0.7039 &0.7054 & 0.7068 & 0.6983 & 0.7011 \\
		GraphSAGE & 0.7173 & 0.7165  & 0.7234 & 0.7287 & 0.7353&0.7384 & 0.7303 & 0.7265 & 0.7240 \\
		SGCN & 0.6956 & 0.7024  & 0.7021 & 0.7098 & 0.7063&0.7135 & 0.7129 & 0.7082 & 0.7076 \\
		GAT & 07105 & 0.7130  & 0.7216 & 0.7242 & 0.7209&0.7287 & 0.7315 & 0.7262 & 0.7099 \\
		GIN & 0.7314 & 0.7268  & 0.7335 & 0.7415 & 0.7388&0.7392 & 0.7426 & 0.7443 & 0.7381 \\
		JK-Net & 0.7453 & 0.7439  & 0.7414 & 0.7576 & 0.7602&0.7618 & 0.7622 & 0.7581 & 0.7590 \\
		\hline
		Graphormer & 0.7505 & 0.7512  & 0.7563 & 0.7625 & 0.7633&0.7647 & 0.7686 & 0.7679 & 0.7624 \\
		SAT & 0.7746 & 0.7809  & 0.7785 & 0.7798 & 0.7859&0.7890 & 0.7987 & 0.7924 & 0.7892\\
		TransGNN & 0.7838 & 0.7843  & 0.7827 & 0.7824 & 0.7836&0.7916 & 0.7952 & 0.7901 & 0.7855\\
		SGFormer & 0.8009 & 0.7948  & 0.8052 & 0.8099 & 0.8140&0.8119 & 0.8078 & 0.8034 & 0.8013\\
		DeepGraph & 0.7954 & 0.7965  & 0.8019 & 0.8087 & 0.8112&0.8163 & 0.8206 & 0.8171 & 0.8180\\
		NAGphormer & 0.8013 & 0.8025  & 0.8048 & 0.8143 & 0.8189&0.8240 & 0.8243 & 0.8174 & 0.8236\\
		TANGNN-FLC & 0.8430 & 0.8405  & 0.8462 & 0.8517 & 0.8493&0.8503 & 0.8566 & 0.8567 & 0.8531\\
		TANGNN-NAI & 0.8414& 0.8421  & 0.8440 & 0.8492 & 0.8508&0.8517 & 0.8589 & 0.8545 & 0.8529\\
		TANGNN-TAI & 0.8423 & 0.8414  & 0.8465 & 0.8536 & 0.8449&0.8536 & 0.8567 & 0.8552 & 0.8568\\
		TANGNN & 0.8508 & \textbf{0.8492}  & 0.8480 & 0.8521 & 0.8559&0.8588& 0.8607 & 0.8634 & 0.8623\\
		TANGNN-LC & \textbf{0.8552} & 0.8487  & \textbf{0.8559} & \textbf{0.8570}& \textbf{0.8628}&\textbf{0.8651} & \textbf{0.8673} & \textbf{0.8711} & \textbf{0.8695}\\
		\bottomrule
	\end{tabularx}
\end{table}

\begin{table}[h]
	\centering
	\setlength{\tabcolsep}{1.9pt}
	\caption{The f1\_micro values obtained by completing the vertex classification task on the Pumbed dataset. The best result is marked in bold.}\label{tab5}%
	\footnotesize 
	\begin{tabularx}{\linewidth}{cccccccccc}\toprule
		Algorithm & 10\% & 20\% & 30\% & 40\% & 50\% & 60\%& 70\%& 80\% & 90\%\\ \hline
		GCN &  0.6283 & 0.6301 & 0.6274 & 0.6295 & 0.6311 &0.6296 & 0.6287 & 0.6275 & 0.6262 \\
		GraphSAGE & 0.6410 & 0.6390  & 0.6422 & 0.6381 & 0.6376&0.6366 & 0.6419 & 0.6375 & 0.6348 \\
		SGCN & 0.6270 & 0.6257 & 0.6278 & 0.6259 & 0.6344&0.6321 & 0.6273 & 0.6266 & 0.6291 \\
		GAT & 0.6406 & 0.6411  & 0.6398 & 0.6417 & 0.6380&0.6377 & 0.6415 & 0.6382 & 0.6373 \\
		GIN & 0.6425 & 0.6414  & 0.6452 & 0.6476 & 0.6397&0.6463 & 0.6411 & 0.6438 & 0.6409 \\
		JK-Net & 0.6503 & 0.6522  & 0.6568 & 0.6554 & 0.6587&0.6571 & 0.6589 & 0.6563 & 0.6520 \\
		\hline
		Graphormer & 0.6584 & 0.6592  & 0.6577 & 0.6633 & 0.6710&0.6698 & 0.6681 & 0.6721& 0.6709\\
		SAT & 0.6813 & 0.6825  & 0.6869 & 0.6944 & 0.6971&0.6962 & 0.7010 & 0.6876 & 0.6848\\
		TransGNN & 0.6725 & 0.6812  & 0.6848 & 0.6892 & 0.6915&0.6874 & 0.6881 & 0.6907 & 0.6810\\
		SGFormer & 0.6937 & 0.6944  & 0.7012& 0.7135 & 0.7230&0.7153 & 0.7098 & 0.7126 & 0.7123\\
		DeepGraph & 0.6993 & 0.6974  & 0.6985 & 0.7203 & 0.7187&0.7195 & 0.7124 & 0.7212 & 0.7186\\
		NAGphormer & 0.7011 & 0.6993  & 0.7025 & 0.7098 & 0.7129&0.7146 & 0.7186 & 0.7200 & 0.7205\\
		TANGNN-FLC & 0.7386 & 0.7428  & 0.7415 & 0.7469 & 0.7456&0.7462 & 0.7470 & 0.7433 & 0.7464\\
		TANGNN-NAI & 0.7412 & 0.7390  & 0.7424 & 0.7426 & 0.7448&0.7495 & 0.7501 & 0.7463 & 0.7475\\
		TANGNN-TAI & 0.7424 & 0.7417  & 0.7462 & 0.7488 & 0.7506&0.7547 & 0.7523 & 0.7511 & 0.7492\\
		TANGNN & 0.7435 & 0.7446  & 0.7509 & 0.7512 & 0.7525&0.7533& 0.7564 & 0.7497 & 0.7510\\
		TANGNN-LC & \textbf{0.7476} & \textbf{0.7523}  & \textbf{0.7565} & \textbf{0.7628}& \textbf{0.7641}&\textbf{0.7696} & \textbf{0.7610} & \textbf{0.7624} & \textbf{0.7585}\\ 
		\bottomrule
	\end{tabularx}
\end{table}

\begin{figure}[h]
	\centering
	\includegraphics[width=\linewidth]{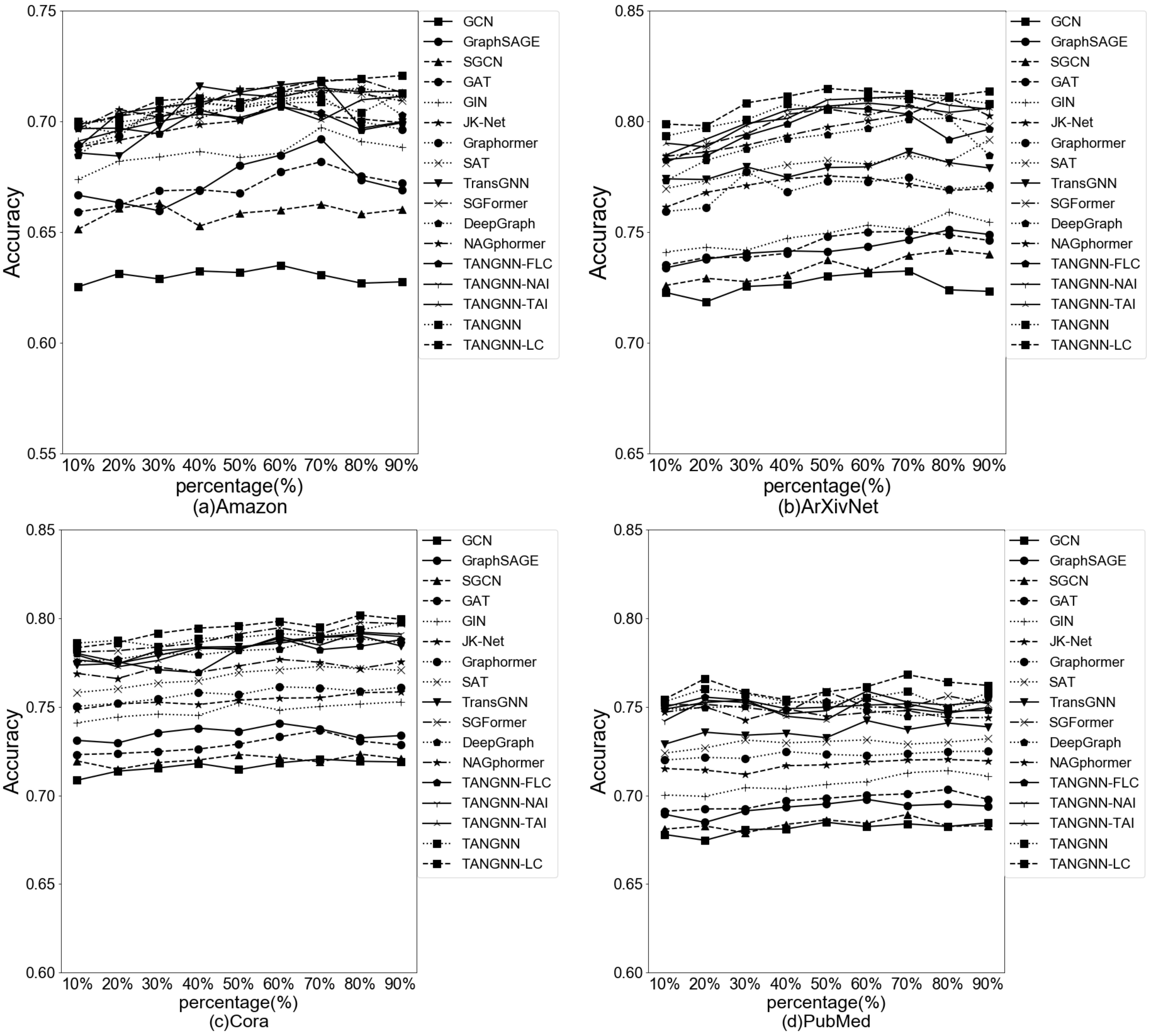}
	\caption{Accuracy values obtained by completing link prediction tasks on four datasets.} 
	\label{fig6}
\end{figure}\textbf{}

\begin{figure}[htbp]
	\centering
	\includegraphics[width=\linewidth]{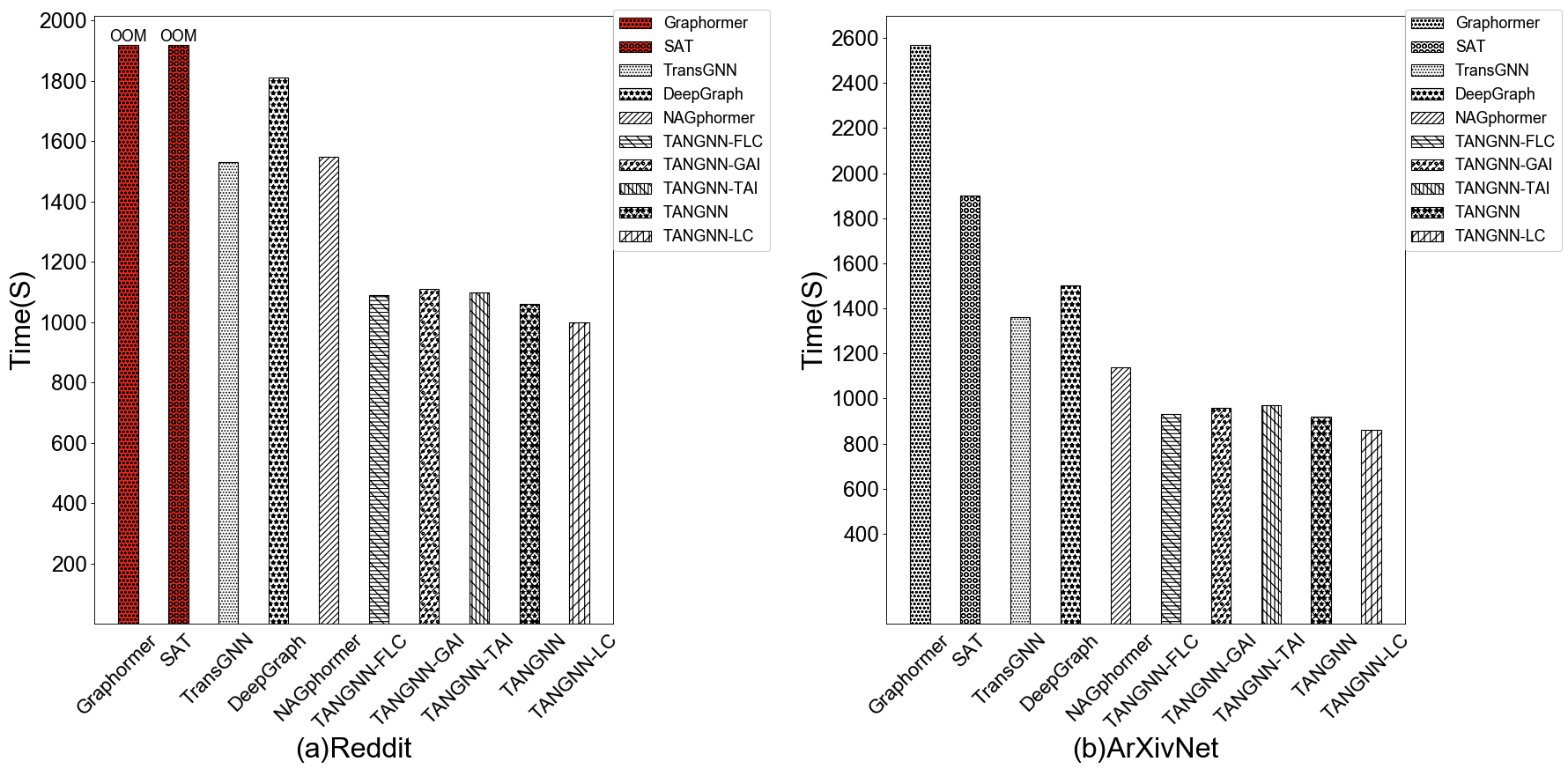}
	\caption{The time spent on link prediction tasks on the ArXivNet and Reddit datasets}
	\label{fig10}
\end{figure}\textbf{}

\subsubsection{LINK PREDICTION}\label{subsubsec10}
The objective of the link prediction task is to forecast potential connections or edges within a network using existing structural information about the network.

Figure~\ref{fig6} shows the results of the link prediction task on the Amazon, ArXivNet, Cora,  PubMed datasets. In the Amazon dataset, As the size of the training set increased, TANGNN demonstrated improved performance. In the ArXivNet dataset, TANGNN-TAI outperformed Graphormer by 6\% (When the training set constitutes 70\% of the total dataset), outperformed TransGNN by about 4\% (When the training set constitutes 70\% of the total dataset), and TANGNN outperformed GraphSAGE by about 9\% (When the training set constitutes 30\% of the total dataset), while TANGNN-LC outperformed GraphSAGE by about 12\% (When the training set constitutes 30\% of the total dataset). In both the Cora and PubMed datasets, TANGNN-LC achieved better performance.

Figure~\ref{fig10} shows our analysis of the runtime for the link prediction task on the ArXivNet and Reddit datasets. In this experiment, the runtime is calculated based on the model reaching a predefined convergence condition rather than a fixed number of iterations, i.e., the training loss of the model reached a stable state and no longer changed significantly. From the figure, it is evident that Graphormer and SAT encountered memory overflow issues on the Reddit dataset, and on the ArXiNet dataset, Graphormer had a longer runtime compared to other algorithms. The experimental results indicate that TANGNN takes significantly less time on large graph datasets compared to TransGNN and NAGphormer, and TANGNN-LC can reach convergence faster than TANGNN, demonstrating its efficiency advantage in processing large-scale graph data.

\subsubsection{Citation Sentiment Prediction}\label{subsubsec11}
The purpose of the citation sentiment prediction task in this article is to classify the sentiment polarity of edges in the citation network based on the citation relationships between literature. 
From the table~\ref{tab6}, it can be seen that the performance of TANGNN-LC is superior to all other algorithms. Compared to GraphSAGE, TANGNN has improved performance by 8\% (when the training dataset is 50\%), while TANGNN-LC has improved performance by about 8\% (when the training dataset is 70\%).

\begin{table}[h]
	\centering
	\setlength{\tabcolsep}{1.9pt}
	\caption{The accuracy value obtained by completing sentiment prediction tasks on the ArXivNet dataset. The best result is marked in bold.}\label{tab6}%
	\footnotesize 
	\begin{tabularx}{\linewidth}{cccccccccc}\toprule
		Algorithm & 10\% & 20\% & 30\% & 40\% & 50\% & 60\%& 70\%& 80\% & 90\%\\ \hline
		GCN &  0.8211 & 0.8181 & 0.8180 & 0.8192 & 0.8174 &0.8233 & 0.8198 & 0.8070 & 0.8147 \\
		GraphSAGE & 0.8234 & 0.8224  & 0.8311 & 0.8300 & 0.8320&0.8307& 0.8277 & 0.8266 & 0.8328 \\
		SGCN & 0.8250 & 0.8210  & 0.8276 & 0.8174 & 0.8334&0.8224 & 0.8237 & 0.8132 & 0.8252 \\
		GAT & 0.8259 & 0.8215  & 0.8296 & 0.8296 & 0.8313&0.8311 & 0.8375 & 0.8281 & 0.8309 \\
		GIN & 0.8320 & 0.8411  & 0.8412 & 0.8372 & 0.8368&0.8366 & 0.8433 & 0.8401 & 0.8411 \\
		JK-Net & 0.8492 & 0.8357  & 0.8565 & 0.8410 & 0.8580&0.8452 & 0.8437 & 0.8450 & 0.8489 \\
		\hline
		Graphormer & 0.8461 & 0.8473  & 0.8521 & 0.8581 & 0.8522&0.8601 & 0.8590 & 0.8528 & 0.8573  \\
		SAT & 0.8546 & 0.8589  & 0.8558 & 0.8649 & 0.8633&0.8737 & 0.8667 & 0.8702 & 0.8696\\
		TransGNN & 0.8609 & 0.8592  & 0.8675 & 0.8627 & 0.8680&0.8722 & 0.8754 & 0.8696 & 0.8717\\
		SGFormer & 0.8674 & 0.8574  & 0.8721 & 0.8633 & 0.8725&0.8752 & 0.8719 & 0.8725 & 0.8718\\
		DeepGraph & 0.8641 & 0.8748  & 0.8690 & 0.8712 & 0.8718&0.8775 & 0.8662& 0.8806 & 0.8676\\
		NAGphormer & 0.8627 & 0.8782  & 0.8713 & 0.8790 & 0.8764&0.8742 & 0.8790 & 0.8713 & 0.8787\\
		TANGNN-FLC & 0.8703 & 0.8653  & 0.8664 & 0.8711 & 0.8714&0.8768 & 0.8809 & 0.8908 & 0.8823\\
		TANGNN-NAI & 0.8710 & 0.8679  & 0.8657 & 0.8725 & 0.8735&0.8792 & 0.8790 & 0.8824 & 0.8811\\
		TANGNN-TAI & 0.8612 & 0.8706  & 0.8633 & 0.8736 & 0.8788&0.8806 & 0.8841 & 0.8827 & 0.8810\\
		TANGNN & 0.8744 & 0.8722  & 0.8795 & 0.8839 & 0.8870&0.8930& 0.8914 & 0.8962 & 0.8924\\
		TANGNN-LC & \textbf{0.8757} & \textbf{0.8784}  & \textbf{0.8821} & \textbf{0.8879}& \textbf{0.8893}&\textbf{0.8972} & \textbf{0.8935} & \textbf{0.9005} & \textbf{0.8971}\\
		\bottomrule
	\end{tabularx}
\end{table}

\subsubsection{Graph Regression}\label{subsubsec12}
The purpose of graph regression task is to predict certain numerical attributes in the graph based on known network structure information, such as importance indicators of nodes, physical and chemical properties of molecules, and other continuous variables. The evaluation metric used in the graph regression task is MAE (Mean Absolute Error), and the smaller the value of MAE, the closer the model's prediction is to the true value. 

As shown in Tables~\ref{tab7} and~\ref{tab8}, it can be seen that TANGNN-FLC, TANGNN-NAI, and TANGNN perform very well on both datasets. This indicates that TANGNN and its variants have strong generalization ability and stability in processing graph regression tasks, and can more accurately predict the importance indicators of nodes and the physical and chemical properties of molecules. In contrast, other models show significant fluctuations in performance under different dataset sizes, with lower stability and accuracy compared to the TANGNN model.

\begin{table}[htbp]
	\centering
	\setlength{\tabcolsep}{1.9pt}
	\caption{The MAE values obtained by completing the graph regression task on the ZINC dataset. The best result is marked in bold.}\label{tab7}%
	\footnotesize 
	\begin{tabularx}{\linewidth}{cccccccccc}\toprule
		Algorithm & 10\% & 20\% & 30\% & 40\% & 50\% & 60\%& 70\%& 80\% & 90\%\\ \hline
		GCN &  0.7134 & 0.7068 & 0.7056 & 0.7109 & 0.7048 &0.7037 & 0.6973 & 0.6997 & 0.7003 \\
		GraphSAGE & 0.7092 & 0.7016  & 0.6934 & 0.6870 & 0.6921&0.6834& 0.6749 & 0.6850 & 0.6931 \\
		SGCN  & 0.7095  & 0.7112 & 0.7013 & 0.6932&0.6953 & 0.6926 & 0.6870 & 0.6898& 0.6927 \\
		GAT & 0.7086 & 0.7001  & 0.6847 & 0.6930 & 0.6988&0.6862 & 0.6753 & 0.6810 & 0.6847 \\
		GIN & 0.6950 & 0.6911  & 0.6923 & 0.6878 & 0.6887&0.6792 & 0.6743 & 0.6790 & 0.6801 \\
		JK-Net & 0.7005 & 0.6930  & 0.6995 & 0.6825 & 0.6783 &0.6636 & 0.6692 & 0.6700 & 0.6931 \\
		\hline
		Graphormer & 0.6852 & 0.6833  & 0.6654 & 0.6739 & 0.6642&0.6581 & 0.6683 & 0.6714 & 0.6809 \\
		SAT & 0.6739	&0.6716	&0.6659	&0.6554	&0.6562	&0.6510	&0.6496	&0.6552 &0.6611\\
		TransGNN & 0.6811	&0.6743	&0.6715	&0.6681	&0.6608	&0.6553	&0.6586	&0.6633	&0.6657\\
		SGFormer & 0.6642	&0.6625	&0.6568	&0.6693	&0.6540	&0.6573	&0.6503	&0.6524	&0.6547\\
		DeepGraph & 0.6677	&0.6610	&0.6589	&0.6605	&0.6572	&0.6508	&0.6492	&0.6568	&0.6583\\
		NAGphormer & 0.6638	&0.6551	&0.6562	&0.6590	&0.6463	&0.6477	&0.6489	&0.6597	&0.6579\\
		TANGNN-FLC &0.6762	&0.6785	&0.6743	&0.6696	&0.6678	&0.6621	&0.6596	&0.6610	&0.6598\\
		TANGNN-NAI &0.6724	&0.6708	&0.6724	&0.6683	&0.6615	&0.6548	&0.6511	&0.6488	&0.6507\\
		TANGNN-TAI &0.6694	&0.6671	&0.6620	&0.6568	&0.6564	&0.6525	&0.6452	&0.6477	&0.6543\\
		TANGNN & \textbf{0.6628}&0.6605	&0.6583	&0.6552	&0.6517	&0.6489	&0.6463	&0.6462	&0.6490\\
		TANGNN-LC & 0.6639 & \textbf{0.6577}  & \textbf{0.6542} & \textbf{0.6521}& \textbf{0.6476}&\textbf{0.6457} & \textbf{0.6426} & \textbf{0.6423} & \textbf{0.6462}\\
		\bottomrule
	\end{tabularx}
\end{table}

\begin{table}[htbp]
	\centering
		\setlength{\tabcolsep}{1.9pt}
	\caption{The MAE values obtained by completing the graph regression task on the QM9 dataset. The best result is marked in bold.}\label{tab8}%
	\footnotesize 
	\begin{tabularx}{\linewidth}{cccccccccc}\toprule
	Algorithm & 10\% & 20\% & 30\% & 40\% & 50\% & 60\%& 70\%& 80\% & 90\%\\ \hline
	GCN &  0.1694	&0.1730	&0.1518	&0.1565	&0.1847	&0.1662	&0.1647	&0.1400	&0.1509 \\
	GraphSAGE & 0.1506	&0.1656	&0.1694	&0.1467	&0.1625	&0.1707	&0.1375	&0.1565	&0.1377 \\
	SGCN & 0.1639	&.1684	&0.1583	&0.1486	&0.1749	&01685	&0.1535	&0.1584	&0.1475 \\
	GAT & 0.1618	&0.1648	&0.1643	&0.1591	&0.1664	&0.1642	&0.1524	&0.1589	&0.1555 \\
	GIN & 0.1538	&0.1496	&0.1557	&0.1511	&0.1480	&0.1502	&0.1532	&0.1399	&0.1407\\
	JK-Net & 0.1442	&0.1416	&0.1434	&0.1513	&0.1407	&0.1349	&0.1387	&0.1396	&0.1388\\
	\hline
	Graphormer & 0.1482	&0.1467	&0.1503	&0.1428	&0.1399&	0.1400	&0.1358	&0.1340	&0.1395 \\
	SAT & 0.1405	&0.1388	&0.1422	&0.1353	&0.1437	&0.1341	&0.1288	&0.1312	&0.1354\\
	TransGNN & 0.1410	&0.1392	&0.1383	&0.1360	&0.1425	&0.1338	&0.1290	&0.1360	&0.1380\\
	SGFormer &0.1382	&0.1385	&0.1376	&0.1340	&0.1314	&0.1265	&0.1247	&0.1293	&0.1321\\
	DeepGraph & 0.1430	&0.1412	&0.1403	&0.1286	&0.1298	&0.1310	&0.1243	&0.1284	&0.1246\\
	NAGphormer & 0.1398	&0.1387	&0.1411	&0.1352	&0.1333	&0.1309	&0.1275	&0.1248	&0.1297\\
	TANGNN-FLC & 0.1411	&0.1403	&0.1396	&0.1363	&0.1355	&0.1368	&0.1320	&0.1298	&0.1311\\
	TANGNN-NAI &0.1403	&0.1411	&0.1365	&0.1382	&0.1321	&0.1282	&0.1262	&0.1254	&0.1289\\
	TANGNN-TAI & 0.1397	&0.1383	&0.1354	&0.1348	&0.1309	&0.1272	&0.1285	&0.1268	&0.1274\\
	TANGNN &0.1380	&0.1368	&0.1298	&0.1304	&\textbf{0.1277}	&0.1286	&0.1254	&0.1253	&0.1263\\
	TANGNN-LC & \textbf{0.1376} & \textbf{0.1360}  & \textbf{0.1270} & \textbf{0.1276}& 0.1286&\textbf{0.1245} & \textbf{0.1222} & \textbf{0.1232} & \textbf{0.1217}\\
		\bottomrule
	\end{tabularx}
\end{table}

\begin{figure}[htbp]
	\centering
	\includegraphics[width=\linewidth]{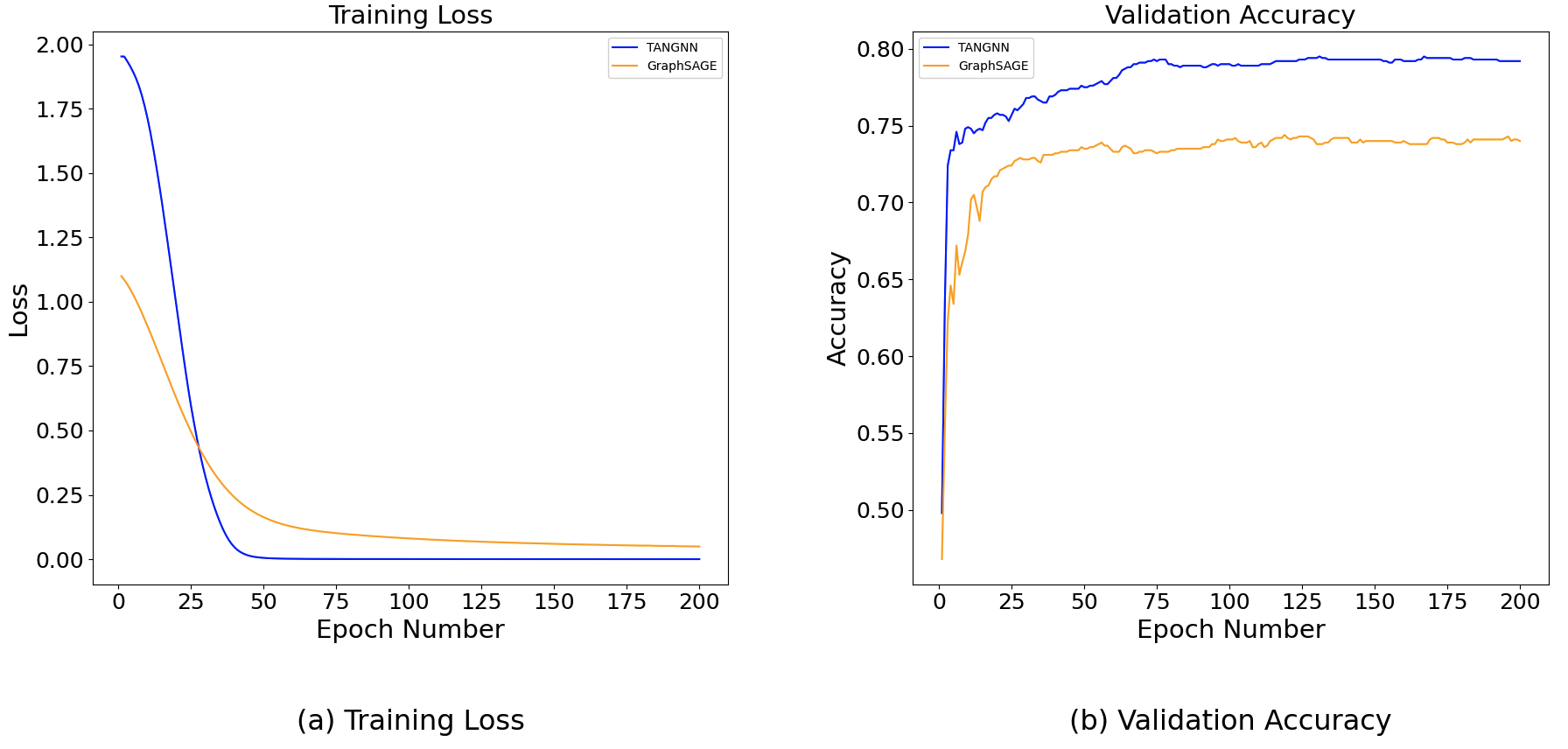}
	\caption{The training process of TANGNN and GraphSAGE}
	\label{fig7}
\end{figure}\textbf{}
\subsubsection{Convergence of TANGNN}\label{subsubsec13}
We investigated the training process of GraphSAGE and TANGNN on the Cora dataset. As shown in Figure~\ref{fig7}, in terms of training loss, TANGNN's loss decreases faster and the final loss is lower, indicating that TANGNN has better fitting effect on the training set and faster convergence speed. In terms of validation accuracy, TANGNN's validation accuracy rapidly increases to a higher level compared to GraphSAGE in the initial stage and maintains a relatively stable state throughout the entire training process.

\begin{figure}[h]
	\centering
	\includegraphics[width=\linewidth]{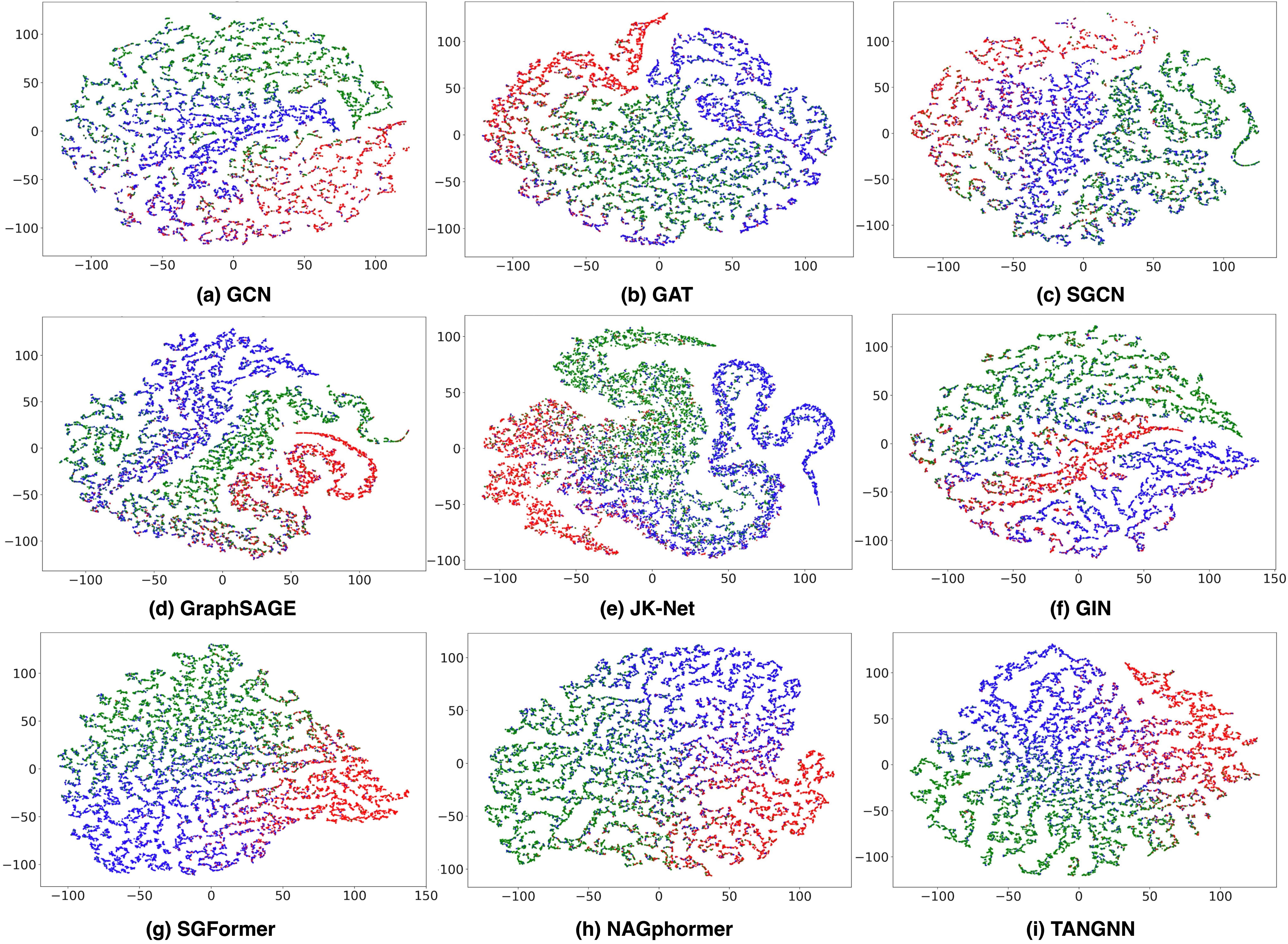}
	\caption{Visualization results on the PubMed dataset. }
	\label{fig8}
\end{figure}\textbf{}
\subsubsection{VISUALIZATION}\label{subsubsec14}

Firstly, we generate representation vectors using various models, which are then visualized in a two-dimensional space using t-SNE~\citep{bib34}. In this visualization, vertices from different classes are distinguished by unique colors. Ideally, vertices belonging to the same class should cluster closely together.

The visualization results, displayed in Figure~\ref{fig8}, utilize the PubMed dataset. The results clearly show effective clustering within small, distinct regions. Our algorithm demonstrates a more distinct separation of points compared to other algorithms, highlighting its effectiveness in distinguishing between classes.
\begin{figure}[htbp]
	\centering
	\includegraphics[width=\linewidth]{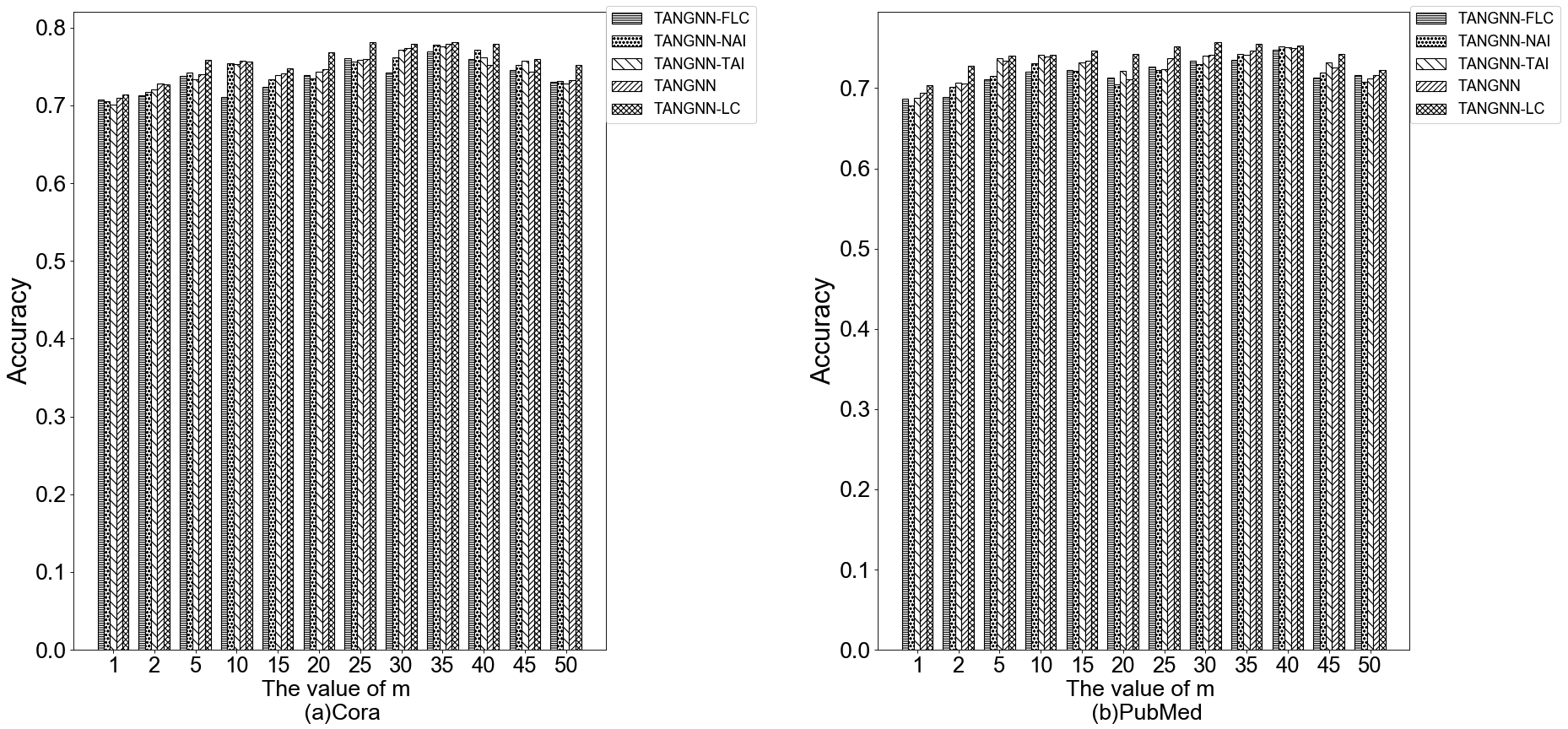}
	\caption{Selection of m for vertex classification tasks on the Cora and PubMed datasets}
	\label{fig9}
\end{figure}\textbf{}
\subsubsection{Parameter Sensitivity Analysis}\label{subsubsec15}
In this study, we evaluated the model's sensitivity to the parameter by varying the number of samples $m$, as shown in Figure~\ref{fig9}. Specifically, we chose a range of sample numbers from 1 to 50, incrementing by 5 each time (for $m$ $\ge$ 5 ), to observe the impact of this variation on the model's accuracy. Our results indicated that on the Cora dataset, the model's accuracy began to decline when the number of samples exceeded 35; on the PubMed dataset, the performance also started to decrease when the number of samples exceeded 40. In conclusion, based on considerations of performance and computational efficiency, we recommend setting the number of samples to 30 to balance model performance and computational costs.

\section{Conclusion}\label{sec13}

In this study, we successfully developed the TANGNN framework, which significantly enhances the efficiency and accuracy of graph neural networks in processing large-scale graph data through an effective integration of the Top-$m$ attention mechanism aggregation and neighbor aggregation components. Specifically, we employed a Top-$m$ sampling strategy for the Top-$m$ attention mechanism aggregation component, and a random neighbor sampling strategy for the neighbor aggregation component, efficiently capturing node information and reducing memory consumption. By concatenating the processed node information from both components at each layer, our model can capture both local and global information within the graph structure, thus adapting to the processing needs of medium to large-scale graph data. Experimental validation showed that TANGNN exhibits outstanding performance on multiple graph processing tasks, proving its potential as a tool for graph data analysis, notably, TANGNN-LC outperformed existing methods on multiple evaluation metrics. Additionally, We developed the ArXivNet dataset by extracting papers from unarXive~\citep{bib40} and constructing a citation network. This study introduces citation sentiment prediction into GNN research for the first time, recorded not only the features of documents but also annotating citation sentiments (positive, neutral, negative).

\section*{CRediT authorship contribution statement}
\textbf{Jiawei E:} Conceptualization, Methodology, Data curation, Software, Writing- original draft, Writing - review \& editing. \textbf{Yinglong Zhang:} Conceptualization, Methodology, Data curation, Supervision, Writing- original draft, Writing - review \& editing. \textbf{Xuewen Xia:} Supervision, Writing - review \& editing. \textbf{Xing Xu:} Supervision, Writing - review \& editing. 

\section*{Declaration of competing interest}
No conflict of interest exits in the submission of this manuscript, and manuscript is approved by all authors for publication. 

\section*{Acknowledgements}
This work was supported by The Natural Science Foundation of Fujian Province of China, No. 2023J01922; Advanced Training Program of Minnan Normal University, No. MSGJB2023015; Headmaster Fund of Minnan Normal University No. KJ19009; Zhangzhou City's Project for Introducing High-level Talents; the National Natural Science Foundation of China, No. 61762036, 62163016.

\bibliographystyle{model5-names} 
\bibliography{mypaper-ref}
	
\end{document}